\pdfoutput=1

\documentclass[11pt]{article}

\usepackage[]{acl}

\usepackage{times}
\usepackage{latexsym}

\usepackage[T1]{fontenc}

\usepackage[utf8]{inputenc}

\usepackage{microtype}
\usepackage{longtable}
\usepackage{inconsolata}
\usepackage{pifont}
\newcommand*\colourcheck[1]{%
  \expandafter\newcommand\csname #1check\endcsname{\textcolor{#1}{\ding{52}}}%
}

\makeatletter
\renewcommand\@makefntext[1]{%
    \parindent 0em%
    \noindent
    \hb@xt@1.8em{\hss\@makefnmark}#1}
\makeatother
\usepackage[hang]{footmisc}

\colourcheck{blue}
\colourcheck{green}
\colourcheck{red}

\usepackage[utf8]{inputenc} 
\usepackage[T1]{fontenc}    
\usepackage{hyperref}       
\usepackage{url}            
\usepackage{booktabs}       
\usepackage{amsfonts}       
\usepackage{nicefrac}       
\usepackage{microtype}      
\usepackage{listings}
\usepackage{algorithm}
\usepackage{algorithmic}
\usepackage{graphicx}
\usepackage{multicol}
\usepackage{CJKutf8}
\usepackage{booktabs}
\usepackage{subcaption}
\usepackage[utf8]{inputenc}
\usepackage{titlesec}
\usepackage[normalem]{ulem}
\usepackage{xspace}
\usepackage{mathtools}
\usepackage{multirow}
\usepackage{graphicx}
\usepackage{subcaption}
\usepackage{pythonhighlight}
\usepackage{wrapfig,lipsum,booktabs}
\usepackage{listings}
\definecolor{codegreen}{rgb}{0,0.6,0}
\definecolor{codegray}{rgb}{0.5,0.5,0.5}
\definecolor{codepurple}{rgb}{0.58,0,0.82}
\definecolor{backcolour}{rgb}{0.95,0.95,0.92}

\lstdefinestyle{mypython}{
    backgroundcolor=\color{backcolour},   
    commentstyle=\color{codegreen},
    keywordstyle=\color{magenta},
    numberstyle=\tiny\color{codegray},
    stringstyle=\color{codepurple},
    basicstyle=\ttfamily\footnotesize,
    breakatwhitespace=false,         
    breaklines=true,                 
    captionpos=b,                    
    keepspaces=true,                 
    numbers=left,                    
    numbersep=5pt,                  
    showspaces=false,                
    showstringspaces=false,
    showtabs=false,                  
    tabsize=2
}

\usepackage[font=small]{caption}
\usepackage{amsthm}
\usepackage{tabularray}
\usepackage[nameinlink]{cleveref}
\crefformat{section}{\S#2#1#3} 
\crefformat{appendix}{\S#2#1#3} 
\crefname{algorithm}{Alg.}{Algs.}
\crefformat{subsection}{\S#2#1#3}
\Crefname{equation}{Eq.}{Eqs.}
\Crefname{figure}{Fig.}{Figs.}
\usepackage[colorinlistoftodos,prependcaption,textsize=tiny]{todonotes}

\usepackage{caption}
\usepackage{soul}

\usepackage{float}
\usepackage{array}

\usepackage{multirow}
\usepackage{hhline}
\usepackage{microtype}
\usepackage{tcolorbox}

\usepackage{inconsolata}

\usepackage{microtype}
\usepackage{hyperref}
\usepackage{booktabs}
\usepackage{graphicx}
\usepackage{footmisc}
\usepackage{subcaption,siunitx,booktabs}
\usepackage{amsthm}
\usepackage{amsmath}

\usepackage[nameinlink]{cleveref}
\crefformat{section}{\S#2#1#3} 

\definecolor{inc}{RGB}{84,123,71}
\definecolor{dec}{RGB}{219, 48, 122}
\usepackage{inconsolata}
\usepackage{url}            
\usepackage{booktabs}       
\usepackage{amsfonts}       
\usepackage{nicefrac}       
\usepackage{microtype}      

\definecolor{kmycolor}{rgb}{0.858, 0.188, 0.878}

\definecolor{bblue0}{RGB}{255,255,255} 
\definecolor{bblue1}{RGB}{230,245,255} 
\definecolor{bblue2}{RGB}{200,225,255} 
\definecolor{bblue3}{RGB}{185,215,255} 
\definecolor{bblue4}{RGB}{170,205,255} 
\definecolor{bblue5}{RGB}{155,195,255} 
\definecolor{bblue6}{RGB}{140,185,255} 
\definecolor{bblue7}{RGB}{125,175,255} 
\definecolor{bblue8}{RGB}{110,165,255} 
\definecolor{bblue9}{RGB}{95,155,255}  
\definecolor{bblue10}{RGB}{80,145,255}  

\usepackage{inconsolata}

\usepackage{graphicx}
\usepackage{arydshln}
\usepackage{wasysym}
\DeclareUnicodeCharacter{2642}{\male}

\title{What Makes a \emph{Good} {Natural Language} Prompt?}

\author{
Do Xuan Long$^{1,3}$, Duy Dinh$^{1}$\thanks{Equal contribution. Works done during the internship at WING, NUS.}, Ngoc-Hai Nguyen$^{1}$\footnotemark[1], \\ \textbf{Kenji Kawaguchi$^{1}$, Nancy F. Chen$^{3}$, Shafiq Joty$^{2}$, Min-Yen Kan$^{1}$}\\
$^{1}$National University of Singapore, $^{2}$Salesforce AI Research, \\$^{3}$Institute for Infocomm Research (I$^2$R), A*STAR \\
\small{xuanlong.do@u.nus.edu}, \small{\{dinhcongduy131200, haibeo2552001\}@gmail.com}, \\
\small{\{kenji,knmnyn\}@nus.edu.sg}, \small{sjoty@salesforce.com}, \small{nfychen@i2r.a-star.edu.sg}
}

\begin{document}
\maketitle
\begin{abstract}
As large language models (LLMs) have progressed towards more human-like and human--AI communications prevalent, prompting has emerged as a decisive component. However, there is limited conceptual consensus on what exactly quantifies \emph{natural language} prompts. We attempt to address this question by conducting a meta-analysis surveying 150+ prompting-related papers from leading NLP and AI conferences (2022–2025), and blogs. We propose a \emph{property- and human-centric} framework for evaluating prompt quality, encompassing 21 properties categorized into six dimensions. We then examine how existing studies assess their impact on LLMs, revealing their imbalanced support across models and tasks, and substantial research gaps. Further, we analyze correlations among properties in high-quality natural language prompts, deriving prompting recommendations. We then empirically explore multi-property prompt enhancements in reasoning tasks, observing that single-property enhancements often have the greatest impact. Finally, we discover that instruction-tuning on property-enhanced prompts can result in better reasoning models. Our findings establish a foundation for property-centric prompt evaluation and optimization, bridging the gaps between human--AI communication and opening new prompting research directions\footnote{Our codes and data will be made publicly available at \href{https://github.com/dxlong2000/NLPromptEval}{here}.}.
\end{abstract}

\section{Introduction}

Pre-trained LLMs \citep{brown2020language,chowdhery2023palm,openai2022chatgpt,touvron2023llama,team2023gemini,guo2025deepseek}, renowned for their ability to generate human-like text, have exhibited exceptional performance across various natural language processing tasks. While their effectiveness is profoundly influenced by the quality of \emph{natural language} prompts \citep{sahoo2024systematic}, the art and science of effective prompts remain underexplored. As human--AI interactions become ubiquitous, developing a deeper understanding of these natural language prompts is crucial since they serve as the primary communication interface between humans and AI systems.

Despite the importance of understanding natural language prompts, there remains limited consensus on how to quantify them. Current approaches rely predominantly on \emph{outcome-centric} measurements, such as model-specific performance metrics \citep{deng-etal-2022-rlprompt,lin2024use, shi2024efficient} and iterative trial-and-error testing \citep{pryzant-etal-2023-automatic,do2024prompt} possibly resulting in prompts optimized for machine interpretation rather than human understanding. This can lead to challenges in interpreting and verifying them, potentially introducing adversarial behaviors in LLMs \citep{zou2023universal, zhu2023autodan} and raising concerns about alignment, transparency, overall reliability, and even human--AI communications. 

Several prompting studies \citep{bsharat2023principled,lin2024howtowrite} and guidelines \citep{openai_prompt_engineering,anthropic2025prompt} recently introduce recommendations enhancing certain \emph{properties} of prompts such as ``Specify the desired length of the output''. These \emph{property-centric} recommendations, focusing on prompt quality rather than model performance, offer interpretable strategies and can complement outcome-centric approaches. However, they have key limitations. First, there is no unified or theoretical property-centric framework that abstractly encompasses such practical recommendations, hindering systematic understanding, analysis, and comparison of these strategies. Second, it is unclear whether these recommendations offer universal benefits across models and tasks or are more model- or task-specific. Third, the interactions among these recommendations and their combined effects on model performance remain understudied.

To address these limitations, we present a meta-analysis to systematically study natural language prompts. We survey prompting papers from top NLP and AI conferences in 2022–2025 and blogs written by top-tech companies (see \Cref{appx:survey-paper} for the full list) and 
identify 21 prompt-level properties across six evaluation dimensions offering a novel \emph{property- and human-centric} perspective (\Cref{sec:theoretical-prompt-dimension}). Building on this, we examine how prior studies assess which models and tasks benefit from enhancing each property, uncovering significant imbalanced distributions in the \#papers supporting each property across models and tasks, and research gaps (\Cref{sec:how-properties-impact-model}). Next, we analyze correlations among these properties in a subset of high-quality natural language prompts, deriving practical recommendations for prompt design (\Cref{sec:how-these-properties-correlate}). We then conduct a case study on reasoning tasks to understand the impact of enhancing multiple prompting properties on model performance (\Cref{sec:a-case-study}). Notably, we observe that different prompting properties influence models differently across tasks, and enhancing multiple properties does not always lead to greater improvements; a single property is often the most effective, and fine-tuning models on property-enhanced instructions further boosts such effectiveness. Our contributions are summarized below:
\begin{enumerate}
    \item We introduce a novel property- and human-centric framework for evaluating the quality of natural language prompts, identifying 21 key properties across six evaluation dimensions to shift the focus from outcome-centric to property-centric assessment.
    \vspace{-1mm}
    \item We conduct a meta-analysis of prior studies from 2022–2025 NLP/AI conferences and blogs to investigate how these properties affect model performance, revealing significant research imbalances and gaps.
    \vspace{-1mm}
    \item We analyze correlations among these properties in a curated set of high-quality prompts, deriving practical recommendations to guide effective prompt design.
    \vspace{-1mm}
    \item We study prompting and fine-tuning models for reasoning tasks, finding that optimizing a single prompting property often outperforms combining multiple ones, with effects varying across tasks and models.

\end{enumerate}

\section{Related work} \label{sec:related-work}

\paragraph{Prompt analysis.} 
Prompting plays a key role in harnessing the full potential of LLMs \citep{liu2023pre,sahoo2024systematic}, driving significant prompt analysis research interest. Existing studies primarily focus on two key directions. The first analyzes the structural components of prompts, highlighting how their variants in terms of formatting \citep{long2024llms} and phrasing \citep{yin-etal-2023-read} can lead to substantial performance differences, and their appearance rates \citep{ma-etal-2024-death}. These studies aim to understand prompt components and their impact on model performance.
The second analyzes prompts through practical experiments, providing design recommendations such as chain-of-thought prompting \citep{wei2022chain, kojima2022large}, being polite with LLMs \citep{ bsharat2023principled} 
and even sets of general guidelines \citep{anthropic2025prompt, openai_prompt_engineering}. 
However, these prompt analysis studies are often task-specific or focus on particular properties of prompts. In this work, for the first time, we introduce a unified property-centric framework that abstractly composites these practical recommendations, facilitating systematic understanding, analysis, and comparison of prompting strategies.

\paragraph{Prompt engineering and optimization.} 
Prompt engineering \citep{wei2022chain,zhangautomatic,zhou2023large} and optimization \citep{deng-etal-2022-rlprompt,pryzant-etal-2023-automatic,do2024prompt} aim to find prompts that maximize a language model's performance for a given task. While much of the existing research focuses on enhancing benchmark performance, there are emerging recent efforts emphasizing broader prompt properties such as clarity \cite{lin2024howtowrite, anthropic2025prompt}, politeness \citep{ bsharat2023principled, yin-etal-2024-respect}, structured formatting \citep{openai_prompt_engineering}, and even fairness in output generation \citep{ji-etal-2023-towards, yuan-etal-2023-causality}. However, it remains unclear whether these properties yield universal benefits across models and tasks or if their effects are model- or task-specific. Furthermore, their interactions and combined influence on model performance remain largely unexplored. We address these gaps in \Cref{sec:how-properties-impact-model,sec:how-these-properties-correlate,sec:a-case-study}.

\begin{table*}
\centering
\footnotesize
\scalebox{.63}{
\begin{tabular}{lllllll}
\toprule
\textbf{Property} & \textbf{Real-world chat} & \textbf{Eval. suit} & \textbf{Reasoning/QA} & \textbf{Generation} & \textbf{NLU} & \textbf{Others} \\
 & AlpacaEval/ATLAS/ & MMLU/C-Eval/ & GSM8K/Comm.QA/ & CNN/Arxiv-March23/ & GLUE/CommitmentBank/ & Safety/Persona./ \\
 & ShareGPT/\dots & BIG-Bench/\dots & HotpotQA/ELI5/\dots & HumanEval/Translation/\dots & DBPedia/\dots & Judging/Retrieval/\dots \\
\midrule
Better quantity & \cellcolor{bblue4}4 \includegraphics[width=0.35cm]{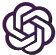} \includegraphics[width=0.35cm]{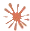} \includegraphics[width=0.35cm]{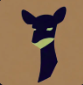} \includegraphics[width=0.35cm]{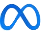} \includegraphics[width=0.35cm]{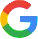}& \cellcolor{bblue4}4 \includegraphics[width=0.35cm]{imgs/chatgpt.png} \includegraphics[width=0.35cm]{imgs/claude.png} \includegraphics[width=0.35cm]{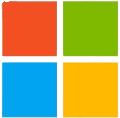} \includegraphics[width=0.35cm]{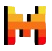}& \cellcolor{bblue9}9 \includegraphics[width=0.35cm]{imgs/chatgpt.png} \includegraphics[width=0.35cm]{imgs/claude.png} \includegraphics[width=0.35cm]{imgs/vicuna.png} \includegraphics[width=0.35cm]{imgs/lllama.png} \includegraphics[width=0.35cm]{imgs/flan.png} & \cellcolor{bblue4}4 \includegraphics[width=0.35cm]{imgs/chatgpt.png} \includegraphics[width=0.35cm]{imgs/claude.png} \includegraphics[width=0.35cm]{imgs/vicuna.png} \includegraphics[width=0.35cm]{imgs/lllama.png} & \cellcolor{bblue1}1  \includegraphics[width=0.35cm]{imgs/chatgpt.png} \includegraphics[width=0.35cm]{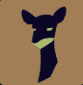} \includegraphics[width=0.35cm]{imgs/lllama.png}& \cellcolor{bblue0}0 \\
 Better manner & \cellcolor{bblue0}0  & \cellcolor{bblue0}0 & \cellcolor{bblue0}0 & \cellcolor{bblue0}0  & \cellcolor{bblue0}0 & \cellcolor{bblue0}0 \\
Better engagement & \cellcolor{bblue2}2 \includegraphics[width=0.35cm]{imgs/chatgpt.png} \includegraphics[width=0.35cm]{imgs/lllama.png}& \cellcolor{bblue0}0 & \cellcolor{bblue1}1 \includegraphics[width=0.35cm]{imgs/chatgpt.png} \includegraphics[width=0.35cm]{imgs/vicuna.png} & \cellcolor{bblue2}2 \includegraphics[width=0.35cm]{imgs/chatgpt.png} \includegraphics[width=0.35cm]{imgs/lllama.png}  & \cellcolor{bblue0}0 & \cellcolor{bblue1}1 \includegraphics[width=0.35cm]{imgs/chatgpt.png} \includegraphics[width=0.35cm]{imgs/lllama.png}  \\
Better politeness & \cellcolor{bblue1}1 \includegraphics[width=0.35cm]{imgs/chatgpt.png} \includegraphics[width=0.35cm]{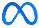}& \cellcolor{bblue2}2 \includegraphics[width=0.35cm]{imgs/chatgpt.png} \includegraphics[width=0.35cm]{imgs/lllama.png} \includegraphics[width=0.35cm]{imgs/mistral.png} \includegraphics[width=0.35cm]{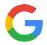} \includegraphics[width=0.35cm]{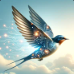} \includegraphics[width=0.35cm]{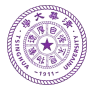} & \cellcolor{bblue1}1 \includegraphics[width=0.35cm]{imgs/chatgpt.png} \includegraphics[width=0.35cm]{imgs/lllama.png} \includegraphics[width=0.35cm]{imgs/swallow.png} & \cellcolor{bblue4}4 \includegraphics[width=0.35cm]{imgs/chatgpt.png} \includegraphics[width=0.35cm]{imgs/lllama.png} \includegraphics[width=0.35cm]{imgs/swallow.png}  \includegraphics[width=0.35cm]{imgs/mistral.png} & \cellcolor{bblue2}2 \includegraphics[width=0.35cm]{imgs/chatgpt.png}
 \includegraphics[width=0.35cm]{imgs/roberta.png}& \cellcolor{bblue2}2 \includegraphics[width=0.35cm]{imgs/lllama.png} \includegraphics[width=0.35cm]{imgs/mistral.png} \includegraphics[width=0.35cm]{imgs/gemini.png} \includegraphics[width=0.35cm]{imgs/chatgpt.png} \\
\midrule
Better intrinsic & \cellcolor{bblue3}3 \includegraphics[width=0.35cm]{imgs/lllama.png} \includegraphics[width=0.35cm]{imgs/chatgpt.png} \includegraphics[width=0.35cm]{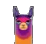} \includegraphics[width=0.35cm]{imgs/vicuna.png}& \cellcolor{bblue2}2 \includegraphics[width=0.35cm]{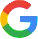} \includegraphics[width=0.35cm]{imgs/chatgpt.png} \includegraphics[width=0.35cm]{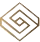} \includegraphics[width=0.35cm]{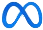} & \cellcolor{bblue7}7 \includegraphics[width=0.35cm]{imgs/chatgpt.png} \includegraphics[width=0.35cm]{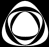} \includegraphics[width=0.35cm]{imgs/opt.png} \includegraphics[width=0.35cm]{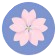} \includegraphics[width=0.35cm]{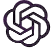} \includegraphics[width=0.35cm]{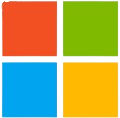} \includegraphics[width=0.35cm]{imgs/chatglm.png}& \cellcolor{bblue2}2 \includegraphics[width=0.35cm]{imgs/chatgpt.png} \includegraphics[width=0.35cm]{imgs/deepseek.png} \includegraphics[width=0.35cm]{imgs/chatglm.png} \includegraphics[width=0.35cm]{imgs/alpaca.png} \includegraphics[width=0.35cm]{imgs/vicuna.png} & \cellcolor{bblue3}3 \includegraphics[width=0.35cm]{imgs/EleutherAI.png} \includegraphics[width=0.35cm]{imgs/opt.png} \includegraphics[width=0.35cm]{imgs/T0.png} \includegraphics[width=0.35cm]{imgs/chatgpt.png} \includegraphics[width=0.35cm]{imgs/chatglm.png} \includegraphics[width=0.35cm]{imgs/alpaca.png} \includegraphics[width=0.35cm]{imgs/vicuna.png}& 
\cellcolor{bblue8}8 \includegraphics[width=0.35cm]{imgs/chatgpt.png} 
\includegraphics[width=0.35cm]{imgs/chatglm.png} \includegraphics[width=0.35cm]{imgs/alpaca.png} 
\includegraphics[width=0.35cm]{imgs/lllama.png} \includegraphics[width=0.35cm]{imgs/vicuna.png} 
\includegraphics[width=0.35cm]{imgs/mistral.png} \includegraphics[width=0.35cm]{imgs/flan.png}\\
Lower extraneous & \cellcolor{bblue0}0 & \cellcolor{bblue1}1 \includegraphics[width=0.35cm]{imgs/chatgpt.png} \includegraphics[width=0.35cm]{imgs/lllama.png} \includegraphics[width=0.35cm]{imgs/mistral.png}& 
\cellcolor{bblue3}3 \includegraphics[width=0.35cm]{imgs/chatgpt.png} \includegraphics[width=0.35cm]{imgs/lllama.png} \includegraphics[width=0.35cm]{imgs/mistral.png} \includegraphics[width=0.35cm]{imgs/longchat.png} \includegraphics[width=0.35cm]{imgs/claude.png}& \cellcolor{bblue0}0 & \cellcolor{bblue0}0 & \cellcolor{bblue3}3 \includegraphics[width=0.35cm]{imgs/chatgpt.png} \includegraphics[width=0.35cm]{imgs/lllama.png} \includegraphics[width=0.35cm]{imgs/vicuna.png}  \includegraphics[width=0.35cm]{imgs/claude.png}\\
Better germane & \cellcolor{bblue1}1  \includegraphics[width=0.35cm]{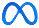} \includegraphics[width=0.35cm]{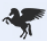} & \cellcolor{bblue1}1 \includegraphics[width=0.35cm]{imgs/chatgpt.png}  & \cellcolor{bblue2}2 \includegraphics[width=0.35cm]{imgs/chatgpt.png} \includegraphics[width=0.35cm]{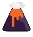} & \cellcolor{bblue1}1 \includegraphics[width=0.35cm]{imgs/bart.png} \includegraphics[width=0.35cm]{imgs/pegasus.png} & \cellcolor{bblue0}0 & \cellcolor{bblue0}0 \\
\midrule
Better objective(s) & \cellcolor{bblue1}1 \includegraphics[width=0.35cm]{imgs/chatgpt.png} \includegraphics[width=0.35cm]{imgs/lllama.png} & \cellcolor{bblue1}1 \includegraphics[width=0.35cm]{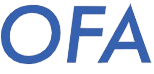} \includegraphics[width=0.35cm]{imgs/mDeberta.png} & \cellcolor{bblue1}1 \includegraphics[width=0.5cm]{imgs/ofa.png} \includegraphics[width=0.35cm]{imgs/mDeberta.png} & \cellcolor{bblue1}1 \includegraphics[width=0.35cm]{imgs/chatgpt.png} \includegraphics[width=0.35cm]{imgs/lllama.png} \includegraphics[width=0.35cm]{imgs/mistral.png} \includegraphics[width=0.35cm]{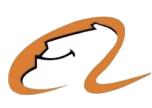} & \cellcolor{bblue1}1 \includegraphics[width=0.5cm]{imgs/ofa.png} \includegraphics[width=0.35cm]{imgs/mDeberta.png} & \cellcolor{bblue0}0 \\
Better external tool(s) & \cellcolor{bblue1}1 \includegraphics[width=0.35cm]{imgs/chatgpt.png} \includegraphics[width=0.35cm]{imgs/alpaca.png} \includegraphics[width=0.35cm]{imgs/vicuna.png} & \cellcolor{bblue2}2 \includegraphics[width=0.35cm]{imgs/chatgpt.png} & \cellcolor{bblue2}2 \includegraphics[width=0.35cm]{imgs/chatgpt.png} \includegraphics[width=0.35cm]{imgs/lllama.png} \includegraphics[width=0.35cm]{imgs/palm.png}& \cellcolor{bblue1}1 \includegraphics[width=0.35cm]{imgs/chatgpt.png} & \cellcolor{bblue0}0 & \cellcolor{bblue1}1 \includegraphics[width=0.35cm]{imgs/chatgpt.png} \includegraphics[width=0.35cm]{imgs/lllama.png} \\
Better metacognition & \cellcolor{bblue0}0 & \cellcolor{bblue2}2 \includegraphics[width=0.35cm]{imgs/chatgpt.png} \includegraphics[width=0.35cm]{imgs/lllama.png} \includegraphics[width=0.35cm]{imgs/claude.png}& \cellcolor{bblue2}2 \includegraphics[width=0.35cm]{imgs/chatgpt.png} \includegraphics[width=0.35cm]{imgs/lllama.png} \includegraphics[width=0.35cm]{imgs/palm.png} & \cellcolor{bblue0}0 & \cellcolor{bblue1}1 \includegraphics[width=0.35cm]{imgs/chatgpt.png} \includegraphics[width=0.35cm]{imgs/lllama.png} \includegraphics[width=0.35cm]{imgs/palm.png} & \cellcolor{bblue1}1 \includegraphics[width=0.35cm]{imgs/chatgpt.png} \includegraphics[width=0.35cm]{imgs/lllama.png} \includegraphics[width=0.35cm]{imgs/claude.png}\\
Better demo(s) & \cellcolor{bblue1}1 \includegraphics[width=0.35cm]{imgs/lllama.png} \includegraphics[width=0.35cm]{imgs/chatgpt.png} & \cellcolor{bblue2}2 \includegraphics[width=0.35cm]{imgs/palm.png} 
\includegraphics[width=0.35cm]{imgs/vicuna.png} \includegraphics[width=0.35cm]{imgs/chatgpt.png} & \cellcolor{bblue8}8 \includegraphics[width=0.35cm]{imgs/chatgpt.png} \includegraphics[width=0.35cm]{imgs/vicuna.png}
\includegraphics[width=0.35cm]{imgs/palm.png} \includegraphics[width=0.35cm]{imgs/T0.png} \includegraphics[width=0.35cm]{imgs/lllama.png} \includegraphics[width=0.35cm]{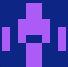} \includegraphics[width=0.35cm]{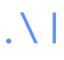} & \cellcolor{bblue4}4 \includegraphics[width=0.35cm]{imgs/lllama.png} \includegraphics[width=0.35cm]{imgs/chatgpt.png} \includegraphics[width=0.35cm]{imgs/vicuna.png}  & \cellcolor{bblue3}3 \includegraphics[width=0.35cm]{imgs/lllama.png} \includegraphics[width=0.35cm]{imgs/chatgpt.png}  \includegraphics[width=0.35cm]{imgs/palm.png} & \cellcolor{bblue1}1  \includegraphics[width=0.35cm]{imgs/palm.png}  \includegraphics[width=0.35cm]{imgs/T0.png}\\  
Better reward(s) & \cellcolor{bblue1}1 \includegraphics[width=0.35cm]{imgs/lllama.png} \includegraphics[width=0.35cm]{imgs/chatgpt.png} & \cellcolor{bblue2}2 \includegraphics[width=0.35cm]{imgs/lllama.png}  & \cellcolor{bblue2}2 \includegraphics[width=0.35cm]{imgs/lllama.png}  & \cellcolor{bblue1}1  \includegraphics[width=0.35cm]{imgs/chatgpt.png} & \cellcolor{bblue0}0 & \cellcolor{bblue1}1 \includegraphics[width=0.35cm]{imgs/lllama.png} \\
\midrule
Better structure & \cellcolor{bblue1}1 \includegraphics[width=0.35cm]{imgs/lllama.png} \includegraphics[width=0.35cm]{imgs/chatgpt.png} & \cellcolor{bblue1}1 \includegraphics[width=0.35cm]{imgs/qwen.png} \includegraphics[width=0.35cm]{imgs/deepseek.png} \includegraphics[width=0.35cm]{imgs/lllama.png} \includegraphics[width=0.35cm]{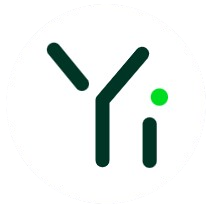} \includegraphics[width=0.35cm]{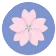} \includegraphics[width=0.5cm]{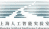}& \cellcolor{bblue4}4 \includegraphics[width=0.35cm]{imgs/qwen.png} \includegraphics[width=0.35cm]{imgs/deepseek.png} \includegraphics[width=0.35cm]{imgs/lllama.png} \includegraphics[width=0.35cm]{imgs/yi.png} \includegraphics[width=0.35cm]{imgs/bloom.png} \includegraphics[width=0.5cm]{imgs/internlm.png} \includegraphics[width=0.35cm]{imgs/chatgpt.png} \includegraphics[width=0.35cm]{imgs/palm.png}& \cellcolor{bblue2}2 \includegraphics[width=0.35cm]{imgs/gemini.png} \includegraphics[width=0.35cm]{imgs/chatgpt.png} & \cellcolor{bblue1}1  \includegraphics[width=0.35cm]{imgs/chatgpt.png} & \cellcolor{bblue0}0 \\
Better context logic & \cellcolor{bblue0}0 & \cellcolor{bblue0}0 & \cellcolor{bblue1}1 \includegraphics[width=0.35cm]{imgs/chatgpt.png} & \cellcolor{bblue0}0 & \cellcolor{bblue0}0 & \cellcolor{bblue1}1 \includegraphics[width=0.35cm]{imgs/mistral.png} \includegraphics[width=0.35cm]{imgs/qwen.png} \includegraphics[width=0.35cm]{imgs/chatgpt.png} \includegraphics[width=0.35cm]{imgs/lllama.png} \includegraphics[width=0.35cm]{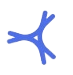}\\
\midrule
Better hallu. awa. & \cellcolor{bblue0}0  & \cellcolor{bblue0}0 & \cellcolor{bblue1}1 \includegraphics[width=0.35cm]{imgs/chatgpt.png}& \cellcolor{bblue1}1 \includegraphics[width=0.35cm]{imgs/chatgpt.png} \includegraphics[width=0.35cm]{imgs/bart.png} \includegraphics[width=0.35cm]{imgs/pegasus.png}& \cellcolor{bblue0}0 & \cellcolor{bblue0}0 \\
Better fact. and cre. & \cellcolor{bblue0}0 & \cellcolor{bblue0}0 & \cellcolor{bblue0}0 & \cellcolor{bblue0}0 & \cellcolor{bblue0}0 & \cellcolor{bblue0}0 \\
\midrule
Lower bias & \cellcolor{bblue1}1 \includegraphics[width=0.35cm]{imgs/chatgpt.png} 
\includegraphics[width=0.35cm]{imgs/lllama.png} & \cellcolor{bblue0}0 & \cellcolor{bblue0}0 & \cellcolor{bblue1}1 \includegraphics[width=0.35cm]{imgs/chatgpt.png} \includegraphics[width=0.35cm]{imgs/bart.png}
\includegraphics[width=0.35cm]{imgs/lllama.png}& \cellcolor{bblue0}0 & \cellcolor{bblue2}2 \includegraphics[width=0.35cm]{imgs/chatgpt.png} 
\includegraphics[width=0.35cm]{imgs/vicuna.png} \includegraphics[width=0.35cm]{imgs/claude.png} \includegraphics[width=0.35cm]{imgs/lllama.png} \\
Better safety & \cellcolor{bblue0}0 & \cellcolor{bblue0}0 & \cellcolor{bblue0}0 & \cellcolor{bblue0}0 & \cellcolor{bblue0}0 & \cellcolor{bblue1}1 \includegraphics[width=0.35cm]{imgs/lllama.png} \includegraphics[width=0.35cm]{imgs/vicuna.png} \includegraphics[width=0.35cm]{imgs/mistral.png} \includegraphics[width=0.35cm]{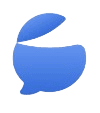} \includegraphics[width=0.35cm]{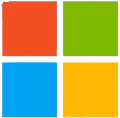}\\
Better privacy & \cellcolor{bblue0}0 & \cellcolor{bblue0}0 & \cellcolor{bblue0}0 & \cellcolor{bblue0}0 & \cellcolor{bblue0}0 & \cellcolor{bblue1}1 \includegraphics[width=0.35cm]{imgs/chatgpt.png} \includegraphics[width=0.35cm]{imgs/lllama.png}\\
Better reliability & \cellcolor{bblue0}0 & \cellcolor{bblue1}1 \includegraphics[width=0.35cm]{imgs/chatgpt.png} \includegraphics[width=0.35cm]{imgs/mistral.png} & \cellcolor{bblue1}1 \includegraphics[width=0.35cm]{imgs/chatgpt.png} \includegraphics[width=0.35cm]{imgs/roberta.png}& \cellcolor{bblue0}0 & \cellcolor{bblue0}0 & \cellcolor{bblue1}1 \includegraphics[width=0.35cm]{imgs/chatgpt.png} \includegraphics[width=0.35cm]{imgs/mistral.png} \\
Better societal norms & \cellcolor{bblue0}0 & \cellcolor{bblue0}0 & \cellcolor{bblue0}0 & \cellcolor{bblue0}0 & \cellcolor{bblue0}0 & \cellcolor{bblue0}0 \\
\bottomrule
\end{tabular}}
\caption{
\small{Summary of the number of papers supporting specific properties across various tasks and models. Model logos are used as follows: \includegraphics[width=0.35cm]{imgs/chatgpt.png}}: ChatGPT / Codex; \includegraphics[width=0.35cm]{imgs/lllama.png}: LLaMa / OPT / RoBERTa / BART; \includegraphics[width=0.45cm]{imgs/qwen.png}: Qwen; \includegraphics[width=0.35cm]{imgs/mistral.png}: Mistral; \includegraphics[width=0.35cm]{imgs/alpaca.png}: Alpaca; \includegraphics[width=0.35cm]{imgs/yi.png}: Yi; \includegraphics[width=0.35cm]{imgs/palm.png}: PaLM / FLAN / Gemma; \includegraphics[width=0.35cm]{imgs/bloom.png}: BLOOM / LongChat / T0; \includegraphics[width=0.35cm]{imgs/chatglm.png}: ChatGLM; \includegraphics[width=0.35cm]{imgs/claude.png}: Claude; \includegraphics[width=0.35cm]{imgs/command_R.png}: Command R; \includegraphics[width=0.35cm]{imgs/deepseek.png}: DeepSeek; \includegraphics[width=0.35cm]{imgs/EleutherAI.png}: EleutherAI; \includegraphics[width=0.35cm]{imgs/internlm.png}: InternLM; \includegraphics[width=0.35cm]{imgs/llava.png}: LLaVa;  \includegraphics[width=0.35cm]{imgs/mDeberta.png}: mDeBERTa / Orca / WizardLM; \includegraphics[width=0.35cm]{imgs/ofa.png}: OFA; \includegraphics[width=0.35cm]{imgs/open_chat.png}: OpenChat; \includegraphics[width=0.35cm]{imgs/pegasus.png}: Pegasus; \includegraphics[width=0.35cm]{imgs/polylm.png}: PolyLM; \includegraphics[width=0.35cm]{imgs/swallow.png}: Swallow; \includegraphics[width=0.35cm]{imgs/vicuna.png}: Vicuna; \includegraphics[width=0.35cm]{imgs/xglm.png}: XGLM. The distribution of papers supporting various properties is highly imbalanced across models and tasks. We discuss the findings in detail in \Cref{sec:how-properties-impact-model}.}
\label{tab:property-impact}
\end{table*}

\section{Prompt quality evaluation} \label{sec:theoretical-prompt-dimension}






We begin our study by conducting a comprehensive survey of over 150 papers and blogs. Our methodology is straightforward: we first examine papers published in ACL, EMNLP, NAACL from ACL Anthology\footnote{\url{https://aclanthology.org/}}, and ICLR, and NeurIPS on OpenReview\footnote{\url{https://openreview.net/}} from 2022 to 2025. Relevant papers are further identified through keyword searches on Google. While striving for thoroughness, we acknowledge the possibility of inadvertently omitting some related papers. We then manually identify prompting objectives and recommendations from these papers that influence model performance, and conceptualize them as prompt \textbf{properties}. These properties are defined below along with its evidence (denoted by abbreviation \textbf{e.b.}).


\paragraph{I. Communication and language.} Prior studies highlight the importance of specific communication properties for desired LM outcomes. For example, \citet{yin-etal-2024-respect} find that impolite prompts degrade model results across tasks and languages, while \citet{shi2023large} discover that irrelevant contexts can distract LLMs, and more explicit prompts enhance model performance \citep{bsharat2023principled, lin2024howtowrite}. Inspired by these and LLMs being more humanoid, prompt evaluation should consider human-like communication properties. We introduce four for evaluation, partially motivated by Grice's Maxims of Conversation \citep{grice1975logic}:

\begin{itemize}
    \item \textbf{Token quantity:} The extent to which prompts provide optimal and relevant information while minimizing token usage, balancing information completeness with efficiency (e.b. \citet{shi2023large,jiang-etal-2023-llmlingua}).
    
    \item  \textbf{Manner:} The degree to which prompts are clear and direct (across turns) while minimizing unnecessary ambiguity, complexity, and confusion (e.b. \citet{anthropic2025prompt}).

    \item \textbf{Interaction and engagement:} The extent to which the prompts explicitly encourage the models to gather the necessary details and requirements by asking questions of clarification or confirmation (e.b. \citet{deng-etal-2023-prompting}).

    \item \textbf{Politeness:} The degree to which prompts maintain respectful, professional, and context-specific politeness, including the use of courteous language (e.g., ``please'', ``thank you'') (e.b. \citet{yin-etal-2024-respect}).
\end{itemize}


\paragraph{II. Cognition.} \label{subsec:cognitive-dim}
\citet{wei2022chain,zhou2023leasttomost} pioneer in introducing prompting methods that decompose complex reasoning tasks into simpler steps, enhancing LLM performance. Subsequent studies extensively investigate strategies that optimize the subtasks to further align them with model capabilities \citep{khot2023decomposed,suzgun2024meta}. In addition, \citet{sun-etal-2022-tsgp} show that integrating self-generated knowledge improves question answering performance of LLMs. Philosophically, these works imply that maximizing LLMs' learning and problem-solving requires meticulous management of their cognitive loads.

\citet{sweller1991evidence} introduce Cognitive Load Theory,  categorizing cognitive loads into {intrinsic} (task complexity), {extraneous} (unclear or poorly designed instructions), and {germane} (efforts to understand, memorize, and organize information). Motivated by this, prompt evaluation should concern three loads on LLMs:

\begin{itemize}
    \item \textbf{Manage intrinsic load:} This evaluates the prompts in explicitly guiding models to break complex tasks into actionable steps aligned with LM skills (e.b. \citet{zhou2023leasttomost}). 
    
    \item \textbf{Reduce extraneous load:} The extent to which prompts minimize unnecessary complexity via simplifying language and removing redundant or irrelevant information to reduce unnecessary load (e.b. \citet{openai_prompt_engineering}). 
    
    \item \textbf{Encourage germane load:} The degree to which prompts explicitly engage models with their prior knowledge or deep working memory (e.g., ``ask itself'' \citep{press-etal-2023-measuring}) to integrate it with existing and new knowledge for problem-solving (e.b. \citet{ sun-etal-2022-tsgp, mialonaugmented,fan2024survey}).

\end{itemize}


\paragraph{III. Instruction.} \label{subsec:instruction-dim}

The instructional values of prompts are crucial for achieving the desired output \citep{sahoo2024systematic}. Drawing on Gagne's Nine Events of Instruction \citep{gagné1985conditions} and the Metacognitive Theories \citep{schraw1995metacognitive}, we present instructional criteria to evaluate them non-overlapping with other dimensions:

\begin{itemize}
    \item \textbf{Objective(s):} How well prompts explicitly communicate the task objectives, including expected personae, outputs, formats, constraints, audiences, and other applicable criteria (e.b. \citet{chang2023prompting,anonymous2024beyond}).

    \item \textbf{External tool(s):} The extent to which prompts explicitly guide models to identify when specific external tools or knowledge resources are needed that go beyond task objective(s), and perform corresponding external calls (e.b. \citet{yao2023react}).  
        
    \item \textbf{Metacognition:} This assesses prompts in explicitly guiding models to reason, self-monitor, and self-verify outputs to meet expectations and enhance reliability (e.b. \citet{wang-zhao-2024-metacognitive}).

    \item \textbf{Demo(s):} The extent to which the prompts explicitly include examples, demonstrations, and counterexamples to illustrate the desired output (e.b. \citet{dong-etal-2024-survey}).
    
    \item \textbf{Reward(s):} How well prompts explicitly establish feedback and reinforcement mechanisms that encourage the models to achieve desired outputs (e.b. \citet{bsharat2023principled}).
\end{itemize}



\paragraph{IV. Logic and structure.} \label{subsec:logic-dim}

Coherent structural prompts are shown to be effective across various tasks  \citep{wang2024langgpt,huang2024can}. Moreover, prompting guidelines \citep{promptingguide2024,openai_prompt_engineering} also recommend structuring input and output to obtain better performing prompts. For logic, recent studies \citep{wang2024resolving, pham-etal-2024-whos} highlight the importance of contextual consistency where knowledge conflicts within prompts substantially degrade LM performance. Building on these insights and the established human logic criteria for effective communication \citep{grice1975logic, Mercier_Sperber_2011}, we introduce two logical criteria:



\begin{itemize}

\item \textbf{Structural logic:} This evaluates the logical clarity and coherence of prompts' structure, and the progression between components (e.b. \citet{wang2024langgpt,zhou2024selfdiscover}).

\item \textbf{Contextual logic:} This assesses the logical consistency and coherence of the instructions, terminologies, concepts, facts, and other components within the prompt and across communication turns (e.b. \citet{pham-etal-2024-whos}).
\end{itemize}


\paragraph{V. Hallucination.} \label{subsec:hallucination-dim}
Prompting can lead to hallucination where models generate plausible but non-factual content \citep{hallucinationllm2024}. While it remains challenging to anticipate whether and when a prompt triggers hallucination \citep{farquhar2024detecting}, prompts can be designed to encourage models to be aware of this critical issue. We propose that prompt evaluation should address two hallucination-related criteria:


\begin{itemize}
\item \textbf{Hallucination awareness:} The extent to which prompts explicitly guide models to generate factual and evidence-based responses while minimizing speculative or unsupported claims (e.b. \citet{gao-etal-2023-enabling}).

\item \textbf{Balancing factuality with creativity:} The degree to which prompts explicitly guide models to balance creative generation with factual accuracy, including which task and when to prioritize creativity over creativity and vice versa. We have yet observed prompting methods designed for this criterion to date. However, \citet{sinha2023mathematical} propose a training approach to balance these aspects for LMs. 
\end{itemize}

In this dimension, we do not evaluate hallucination within prompts as it partially overlaps with the ``Quantity'' of Communication. 

\paragraph{VI. Responsibility.} \label{subsec:ethic-dim}
This dimension emphasizes responsible prompting that mitigates concerns related to inclusion, privacy, safety, bias, reliability, fairness, transparency, and societal norms  \citep{ethicconcerns2024,ethicpromptchatgpt2024}, especially tasks involving sensitive topics or diverse audiences:

\begin{itemize}
    
    \item \textbf{Bias:} The extent to which prompts are devoid of biases and explicitly encourage models to generate content that is free from cultural, gender, racial, or socio-economic biases and avoids stereotypes (e.b. \citet{si2023prompting}).
    
    \item \textbf{Safety:} The degree to which prompts are free from unsafe content and explicitly encourage models to generate safe outputs, avoiding harmful content such as guidance on hazardous activities or weapon creation (e.g., \citet{zou2023universal,zheng2024on}).
    
    \item \textbf{Privacy:} The extent to which prompts do not contain sensitive privacy information and explicitly encourage the models to generate content free of personally sensitive or identifiable information (e.b. \citet{edemacu2024privacy}).
    
    \item \textbf{Reliability:} How well prompts explicitly encourage explicit reasoning processes and attribution, including acknowledgment of model limitations and uncertainties (e.b. \citet{si2023prompting,long-etal-2024-multi-expert}).

    \item \textbf{Societal norms:} The degree to which prompts exclude harmful norms and explicitly encourage models to generate inclusive and appropriate content aligning with widely accepted cultural, ethical, and moral standards (e.b., \citet{yuan-etal-2024-measuring}).

    \end{itemize}

\section{How do properties impact model performance? } \label{sec:how-properties-impact-model}





To assess how the properties in \Cref{sec:theoretical-prompt-dimension} impact model performance, we analyze surveyed papers up to date to determine if these aspects were studied. We categorize the tasks explored into six groups: \emph{(1) Real-world chat}, comprising benchmarks collected from real users such as AlpacaEval \citep{alpaca_eval} and ShareGPT \citep{sharegptShareGPT}; \emph{(2) Evaluation suite}, which have multiple evaluation tasks such as MMLU \citep{hendryckstest2021} and C-Eval \citep{huang2023ceval}; \emph{(3) Reasoning/QA}, covering reasoning and question-answering tasks like GSM8K \citep{cobbe2021training} and HotpotQA \citep{yang-etal-2018-hotpotqa}; \emph{(4) Generation}, focusing on text generation benchmarks such as summarization \citep{nallapati-etal-2016-abstractive}, and translation; \emph{(5) NLU}, encompassing natural language understanding tasks like GLUE \citep{wang-etal-2018-glue} and CommitmentBank \citep{de2019commitmentbank}; and \emph{(6) Others}, which include safety, personalization, judgment, and retrieval tasks. {For each property, we gather three information: the number (\#) of papers supporting the property, tasks that improving the property enhances their performance, and models}.  We discuss our findings in \Cref{tab:property-impact} below as actionable prompting recommendations.

\paragraph{Across tasks.} {There is logical alignment between task requirements and emphasized properties, with notable variations in the \#papers supporting them across tasks.} Firstly, \textbf{in real-world chats, communication properties emerge as the most supported, followed by instruction and cognition properties.} This arises from the practical use of LLMs, where users often craft rich and informative prompts to handle complex and varied tasks. These prompts can extend to tens of thousands of tokens and may sometimes include redundant details \citep{jiang-etal-2023-llmlingua} or lack focus \citep{pan-etal-2024-llmlingua}, particularly in multi-turn interactions \citep{ferron-etal-2023-meep, bsharat2023principled}. Additionally, the significance of instruction properties reflects the interactive nature of chat, while cognition properties are essential for achieving desired outcomes. Secondly, \textbf{for evaluation suites, cognition, instruction, and communication properties are studied the most, with logic additionally emphasized in reasoning/QA tasks.} This aligns with the nature of these benchmarks, where well-cognitive instructions are crucial to strengthen LLM reasoners \citep{wei2022chain, sun-etal-2022-tsgp, qin-etal-2023-cross, bhuiya-etal-2024-seemingly}. Additionally, logic and structure logic also highlight the importance of systematic solving approaches for such tasks \citep{liu2024logic, cheng-etal-2024-structure}.
Thirdly, \textbf{for generation tasks, communication properties receive the most support, followed by the instruction.} This observation reflects the critical importance of efficient token management in generation tasks \citep{jiang-etal-2023-llmlingua, li-etal-2023-compressing, pan-etal-2024-llmlingua}. Interestingly, several studies underscore the effectiveness of incorporating politeness \citep{mishra-etal-2023-pal, xu-etal-2024-llms, mishra-etal-2024-able, yin-etal-2024-respect}, potentially reflecting the inherent biases of LLMs in processing benign rather than informal queries. Fourthly, \textbf{there are limited prompting studies for NLU tasks, and instruction properties appear to be the most explored, followed by cognition properties.} This can be explained by the fact that NLU tasks require models to accurately interpret prompts to reason deeply over language meaning or implications that go beyond surface-level understanding. Finally, \textbf{lower extraneous and better safeguard prompts have been shown to be effective for enhancing safety} \citep{xiao-etal-2024-distract,zheng2024on}; \textbf{better intrinsic for personalization} \citep{lyu-etal-2024-llm,do-etal-2025-aligning}; \textbf{better intrinsic and lower bias for judging} \citep{liu-etal-2023-g,zheng2023judging}; and \textbf{lower extraneous for retrieval} \citep{liu-etal-2024-lost}. While these findings highlight the nuanced alignment between task requirements and the properties shown, significant research gaps remain in exploring how enhancing other properties can further improve model performance on these tasks.








\paragraph{Across models and properties.} 
{We observe that the distribution of model explorations across properties is highly imbalanced}. Specifically, OpenAI's proprietary models (CodeX \citep{chen2021evaluating}, InstructGPT \citep{ouyang2022training}, ChatGPT \citep{openai2022chatgpt}, GPT-4/4o \citep{openai2023gpt4,openai2024gpt4o}) have been the most extensively studied, followed by open-source LLaMa models \citep{touvron2023llama, touvron2023llama2, dubey2024llama}, and Google's models (FLAN \citep{chung2024scaling}, PaLM \citep{chowdhery2023palm}, Gemma \citep{team2024gemma}). {This raises concerns regarding the transferrable effectiveness of these properties across models}. We hypothesize that different properties benefit models differently and that these benefits may also differ across tasks, and validate it in \Cref{sec:a-case-study}.

\textbf{Our analysis reveals task-specific versus universal properties}: while better intrinsic load management, demonstrations, and external tools emerge as being universally effective, hallucination-awareness and responsibility appear to be more task-specific. Better intrinsic load highlights the current LLM weaknesses in implicitly and effectively decomposing complex tasks into more manageable subtasks without explicit guidance. Moreover, demonstration property underscores the value of learning from examples, while using external tools indicates that even with reduced cognitive load and good demonstrations, LLMs still benefit from tools for certain tasks. 

\paragraph{Open questions (Oq).} 
(\textbf{Oq1}) The effectiveness of properties varies across models due to differences in their inherent knowledge, thus, it is an open question whether and when a property beneficial to one model is useful for another. In addition, the missing entries in \Cref{tab:property-impact} highlight several critical yet unexplored properties. For instance, 
(\textbf{Oq2}), while reasoning is fundamental for humans to address tasks \citep{pearl1998graphical}, it is yet studied whether fostering deeper reasoning (improved germane load), reflective behavior (enhanced metacognition), or responsibility can enhance outcomes of LLMs in real-world chat, evaluation suits, and NLU tasks. Moreover, (\textbf{Oq3}), despite creativity's intuitive importance for multiple tasks such as generation, its effectiveness on LLMs remains an open question. Additionally, significant gaps remain in understanding property dynamics, particularly (\textbf{Oq4}) the conditions under which certain relevant or even task-irrelevant properties \citep{taveekitworachai-etal-2024-null} become effective and why. 
Lastly, (\textbf{Oq5}), the observation regarding task-specific and universal properties raises important questions about whether prompt engineering and optimization should prioritize one over the other and which is more significant. Studying (\textbf{Oq1})-(\textbf{Oq5}) holds huge potential for advancing the efficiency, reliability, and alignment of LLMs. Future research could pursue comparative studies across diverse LLMs and tasks, develop quantifiable metrics to evaluate prompts across multiple dimensions, and explore hybrid strategies blending task-specific and universal prompt properties.

\section{How do these properties appear and correlate in high-quality prompts?} \label{sec:how-these-properties-correlate}

We study high-quality natural language prompts to investigate the correlations between these properties to derive prompting recommendations. We manually collect our test set consisting of 765 single-turn prompts from prompt engineering papers, ChatGPT Prompts Collections\footnote{\href{https://ignacio-velasquez.notion.site/4b65ed147bcb499e9f9459c27605d0e7?v=931596b360b24cc4a43eb1788a31407e}{\texttt{ChatGPT Prompts Collections}}}, Awesome ChatGPT Prompts\footnote{\url{https://github.com/f/awesome-chatgpt-prompts}}, Alpaca \citep{alpaca}, Natural Instructions \citep{mishra-etal-2022-cross}, Complex Instructions \citep{he-etal-2024-complex}, and 50 real-world multi-turn ($> 2$ turns) conversations from LMSYS-Chat-1M \citep{zheng2024lmsyschatm} having 204 prompts, totaling 969 prompts in Appx.-\Cref{tab:dataset-statistics}. We evaluate these prompts across 21 proposed properties using GPT-4o-2024-11-20 \citep{openai2024gpt4o} with Self-consistency \citep{wang2022self} as the judge. We also test open-source models, including DeepSeek R1 Distill Qwen 32B \citep{guo2025deepseek} and Mistral Small 24B It 2501 \citep{jiang2023mistral}, as judges. However, we do not use them ultimately since we face significant evaluation format following issues \citep{long2024llms} with DeepSeek and Mistral achieving only 65.42\% and 71.19\%. In addition to GPT-4o, we supplement our correlation results with findings from Gemini-2.0-flash \citep{team2023gemini} in Appendix \Cref{sec:corre_gemini}.

\begin{figure*}
\centering
\includegraphics[width=2\columnwidth, trim={0cm 0cm 0cm 0cm},clip]{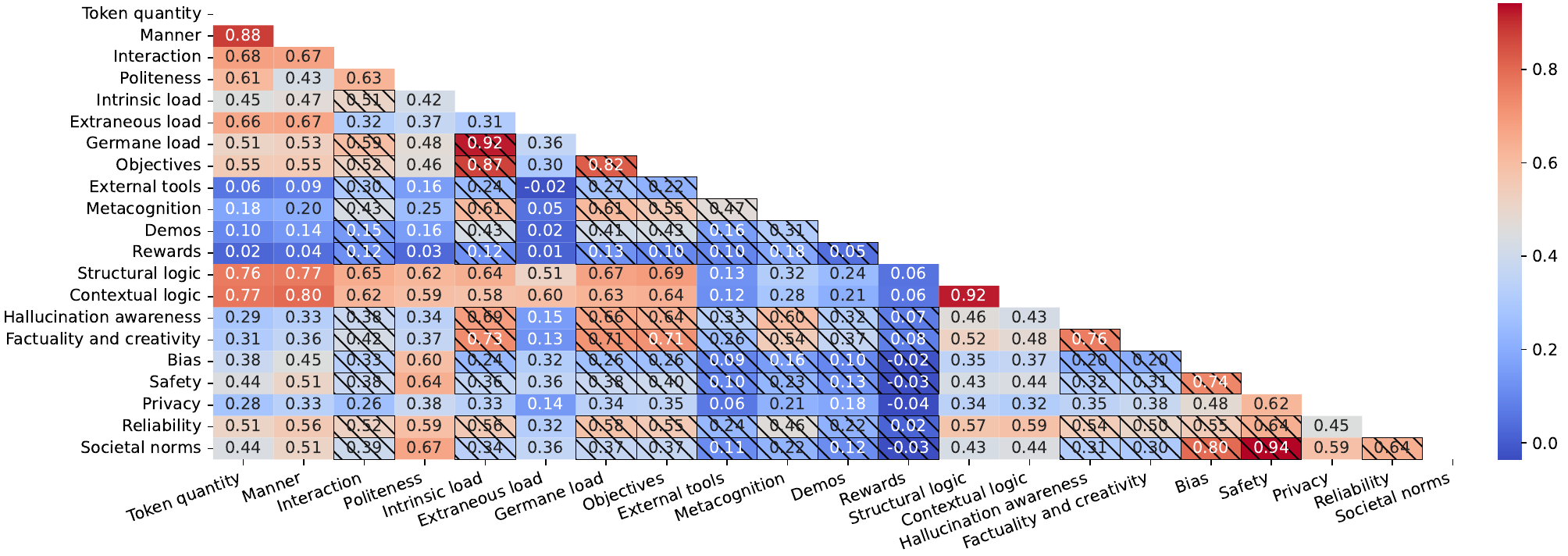}
\vspace{-3mm}
\caption{\small{Correlations of properties evaluated by GPT-4o. We do not consider correlations between pairs of properties concurrently having average scores below 5/10 (hatched by ``\textbackslash\textbackslash'') since they naturally but may falsely suggest correlations.}}
\label{fig:gpt4o-correlation-outcome}
\end{figure*}

\paragraph{Methods.} 
Automatic evaluations using LLMs can be unreliable, especially given the variability in evaluation prompts \citep{doostmohammadi-etal-2024-reliable}. This creates a significant challenge in deriving reliable correlation conclusions from these evaluations. To mitigate this, we first manually label 50 random prompts in 21 properties and then design evaluation prompts to closely align with human judgments. Each annotation is agreed upon by our three prompting researchers with bachelor's degrees and at least six months' experience.

For each evaluation dimension, we begin with a prompt similar to the reference-free judging prompt on a scale of 1-10 proposed by \citet{zheng2023judging}. However, we find that this method results in drastically low Cohen's Kappa agreement \citep{cohen1960coefficient} with human raters; 15/21 topics achieved scores below 0.15, see Appx.-\Cref{fig:cohen-kappa-outcome}, {``Ori. eval.''}. We then supplement an incremental grading system for each criterion, {``Ori. eval. + Inc.''}, similar to \citep{yuan2024selfrewarding}, which significantly enhances agreements. Nevertheless, the germane load, objectives, rewards, and responsibility properties continue to score low. {This is because the evaluator tends to score them higher than human based on implicit instructions rather than explicit cues as expected}. To mitigate this issue,  we explicitly instruct the evaluator to judge explicit signals, resulting in significantly better agreements ({``Ours''} in Appx.-\Cref{fig:cohen-kappa-outcome}). We evaluate all prompts with {``Ours''}.

\paragraph{Findings.}
{For this specific set of prompts}, the property correlations are provided in \Cref{fig:gpt4o-correlation-outcome}. We do not consider correlations between properties if both have an average score below 5/10 (hatched by ``\textbackslash\textbackslash'') because low average scores naturally but may falsely suggest correlations. We observe 17/210 strong correlations ($\geq$ 0.7) among 21 properties. Some of them align with their real-world overlaps. For example, token quantity, manner,  structural logic, contextual logic, and extraneous load reflect the natural correlations between token efficiency, clarity, directness, exclusion of irrelevant details, and logical coherence. Within dimensions, we notice structural logic strongly correlates with contextual logic; hallucination awareness with factuality and creativity; safety with societal norms. Surprisingly, we notice strong correlations between objectives and intrinsic load; objectives and germane load; hallucination awareness and reliability. These can be attributed to the nature of effective human prompting: as we optimize intrinsic and/or germane loads, we tend to articulate objectives more clearly. Similarly, enhancing hallucination awareness inherently contributes to reliability awareness. 

We learn prompting recommendations from the analysis of this set of prompts. Firstly, optimizing prompts for directness, clarity, and conciseness may potentially improve token efficiency, and logical coherence, and reduce extraneous cognitive load. Secondly, clear objectives naturally emerge when prompts are logically structured guiding models to self-monitor their generation or execute tasks step-by-step. Thirdly, explicitly incorporating hallucination awareness in prompts may result in better reliability awareness. Lastly, since these prompts were carefully selected by humans, certain non-obvious correlations, such as those between structural logic, contextual logic, token quantity, and manner, suggest that these properties should be optimized jointly.

\paragraph{Open questions (Oq).} 
While our analysis reveals certain correlations among prompt properties, several open questions remain for future investigation. First, \textbf{(Oq6)} we hypothesize that correlations may vary across different pools of prompts especially those that are task-specific, potentially leading to distinct prompting recommendations. We leave this for future research. Secondly, \textbf{(Oq7)} when two properties exhibit a strong correlation, it remains to be determined whether enhancing prompts in one property causally enhances the other or if these properties merely co-occur within our dataset. 
Finally, \textbf{(Oq8)}, understanding how these correlations influence model performance is critical for advancing prompt optimization methods. The investigation of (\textbf{Oq6})–(\textbf{Oq8}) offers a pathway to optimize LLM prompts by analyzing property correlations and eliminating optimization redundancies. Future work could use causal inference tools, such as structural equation modeling, to distinguish mere co-occurrence from influence, and conduct diverse model- and task-specific experiments to quantify these effects more precisely.



\section{Should we enhance properties of prompts during experiments?} \label{sec:a-case-study}

We perform a preliminary investigation into the impact of combining these properties on the performance of model reasoning. Our experiments are performed under two settings: \emph{prompting} (\Cref{subsec:prompting-experiments}) and (2) \emph{fine-tuning} (\Cref{subsec:finetuning-experiments}), and conducted on the {MMLU} \citep{hendryckstest2021}, {CommonsenseQA} \citep{talmor-etal-2019-commonsenseqa} and {ARC-Challenge} \citep{clark2018think}, and {GSM8K}  datasets.

\subsection{Property-enhanced Prompting} \label{subsec:prompting-experiments}

Our prompting experiments are performed with {Llama-3.1-8B-it} \citep{dubey2024llama}, {Qwen2.5-7B-it} \citep{qwen25}, and {OpenAI o3-mini} \citep{openai_o3_mini} focusing on three dimensions: communication, cognitive loads, and instruction. We exclude demonstrations, objectives, and external tools, as prior work extensively explored these properties. We begin with the zero-shot CoT prompt \citep{kojima2022large} {``Answer the following question step-by-step.''}. We then introduce the following modifications: (1) Add {``Please''} to promote {Politeness}; (2) {``Reflect on your prior knowledge to gain a deeper understanding of the problem before solving it.''} to encourage {Germane load}; (3) {“Self-verify your response thoroughly to ensure each reasoning step is correct.”} to promote {Metacognition}; (4) {``You will be awarded 100 USD for every correct reasoning step.''} to improve the {Rewards}.

\begin{table}
\centering
\resizebox{.47\textwidth}{!}{%
\begin{tabular}{l|l|c|c|c|c}
\toprule
& & \textbf{MMLU} & \textbf{Comm.QA} & \textbf{ARC-C} & \textbf{GSM8K} \\ \hline
& Zero-shot CoT & 65.00 & 76.00 & 81.50 & 82.0 \\ 
\multirow{7}*{\rotatebox{90}{\textbf{Llama-3.1-8B-It}}} 
& + Politeness & 68.00\textcolor[RGB]{0,125,0}{$\uparrow$} & \textbf{83.50}\textcolor[RGB]{0,125,0}{$\uparrow$} & \textbf{84.50}\textcolor[RGB]{0,125,0}{$\uparrow$} & \textbf{87.5}\textcolor[RGB]{0,125,0}{$\uparrow$} \\ 
& + Germane load & 66.00\textcolor[RGB]{0,125,0}{$\uparrow$} & 75.50\textcolor[RGB]{250,0,0}{$\downarrow$} & 82.00\textcolor[RGB]{0,125,0}{$\uparrow$} & 82.0\textcolor[RGB]{250,0,0}{$\downarrow$} \\ 
& + Metacognition & 61.00\textcolor[RGB]{250,0,0}{$\downarrow$} & 81.50\textcolor[RGB]{0,125,0}{$\uparrow$} & 81.00\textcolor[RGB]{250,0,0}{$\downarrow$} & 81.5\textcolor[RGB]{250,0,0}{$\downarrow$} \\ 
& + Rewards & 64.00\textcolor[RGB]{250,0,0}{$\downarrow$} & 80.50\textcolor[RGB]{0,125,0}{$\uparrow$} & 82.00\textcolor[RGB]{0,125,0}{$\uparrow$} & 84.0\textcolor[RGB]{0,125,0}{$\uparrow$} \\ 
& + Pol. + Ger. & 67.00\textcolor[RGB]{0,125,0}{$\uparrow$} & 79.50\textcolor[RGB]{0,125,0}{$\uparrow$} & 80.50\textcolor[RGB]{250,0,0}{$\downarrow$} & 80.5\textcolor[RGB]{250,0,0}{$\downarrow$} \\ 
& + Met. + Rew. & 66.00\textcolor[RGB]{0,125,0}{$\uparrow$} & 80.00\textcolor[RGB]{0,125,0}{$\uparrow$} & 83.50\textcolor[RGB]{0,125,0}{$\uparrow$} & 83.5\textcolor[RGB]{0,125,0}{$\uparrow$} \\ 
& + Pol. + Ger. + Met. & \textbf{69.50}\textcolor[RGB]{0,125,0}{$\uparrow$} & 75.00\textcolor[RGB]{250,0,0}{$\downarrow$} & 82.50\textcolor[RGB]{0,125,0}{$\uparrow$} & 81.5\textcolor[RGB]{250,0,0}{$\downarrow$} \\ 
\midrule
& Zero-shot CoT & 45.50 & 55.00 & 59.50 & 76.5 \\
\multirow{7}*{\rotatebox{90}{\textbf{Qwen-2.5-8B-It}}} 
& + Politeness & 41.00\textcolor[RGB]{250,0,0}{$\downarrow$} & 45.50\textcolor[RGB]{250,0,0}{$\downarrow$} & 54.00\textcolor[RGB]{250,0,0}{$\downarrow$} & 79.0\textcolor[RGB]{0,125,0}{$\uparrow$} \\ 
& + Germane load & 44.50\textcolor[RGB]{250,0,0}{$\downarrow$} & \textbf{56.50}\textcolor[RGB]{0,125,0}{$\uparrow$} & 53.50\textcolor[RGB]{250,0,0}{$\downarrow$} & \textbf{90.0}\textcolor[RGB]{0,125,0}{$\uparrow$} \\ 
& + Metacognition & \textbf{52.50}\textcolor[RGB]{0,125,0}{$\uparrow$} & \textbf{56.50}\textcolor[RGB]{0,125,0}{$\uparrow$} & \textbf{62.00}\textcolor[RGB]{0,125,0}{$\uparrow$} & 83.5\textcolor[RGB]{0,125,0}{$\uparrow$} \\ 
& + Rewards & 40.50\textcolor[RGB]{250,0,0}{$\downarrow$} & 48.00\textcolor[RGB]{250,0,0}{$\downarrow$} & 52.00\textcolor[RGB]{250,0,0}{$\downarrow$} & 66.0\textcolor[RGB]{250,0,0}{$\downarrow$} \\ 
& + Pol. + Ger. & 46.00\textcolor[RGB]{0,125,0}{$\uparrow$} & 54.00\textcolor[RGB]{250,0,0}{$\downarrow$} & 59.00\textcolor[RGB]{250,0,0}{$\downarrow$} & 86.5\textcolor[RGB]{0,125,0}{$\uparrow$} \\ 
& + Met. + Rew. & 41.00\textcolor[RGB]{250,0,0}{$\downarrow$} & 55.50\textcolor[RGB]{0,125,0}{$\uparrow$} & 54.50\textcolor[RGB]{250,0,0}{$\downarrow$} & 88.5\textcolor[RGB]{0,125,0}{$\uparrow$} \\ 
& + Pol. + Ger. + Met. & 46.50\textcolor[RGB]{0,125,0}{$\uparrow$} & 53.50\textcolor[RGB]{250,0,0}{$\downarrow$} & \textbf{62.00}\textcolor[RGB]{0,125,0}{$\uparrow$} & 89.5\textcolor[RGB]{0,125,0}{$\uparrow$} \\ 
\midrule
& Zero-shot CoT & \textbf{92.00} & \textbf{88.50} & 94.50 & \textbf{97.0} \\
\multirow{7}*{\rotatebox{90}{\textbf{o3-mini}}} 
& + Politeness & 88.50\textcolor[RGB]{250,0,0}{$\downarrow$} & 87.00\textcolor[RGB]{250,0,0}{$\downarrow$} & 93.50\textcolor[RGB]{250,0,0}{$\downarrow$} & 96.0\textcolor[RGB]{250,0,0}{$\downarrow$} \\ 
& + Germane load & 88.00\textcolor[RGB]{250,0,0}{$\downarrow$} & 82.00\textcolor[RGB]{250,0,0}{$\downarrow$} & \textbf{95.00}\textcolor[RGB]{0,125,0}{$\uparrow$} & 96.5\textcolor[RGB]{250,0,0}{$\downarrow$} \\ 
& + Metacognition & 90.00\textcolor[RGB]{250,0,0}{$\downarrow$} & 85.00\textcolor[RGB]{250,0,0}{$\downarrow$} & 94.00\textcolor[RGB]{250,0,0}{$\downarrow$} & 95.5\textcolor[RGB]{250,0,0}{$\downarrow$} \\ 
& + Rewards & 89.50\textcolor[RGB]{250,0,0}{$\downarrow$} & 85.50\textcolor[RGB]{250,0,0}{$\downarrow$} & 94.50 & 96.0\textcolor[RGB]{250,0,0}{$\downarrow$} \\ 
& + Pol. + Ger. & 81.00\textcolor[RGB]{250,0,0}{$\downarrow$} & 71.00\textcolor[RGB]{250,0,0}{$\downarrow$} & 88.50\textcolor[RGB]{250,0,0}{$\downarrow$} & \textbf{97.0} \\
\bottomrule
\end{tabular}}
\caption{Performance of models (\%) on various tasks under different configurations. Arrows indicate changes relative to Zero-shot CoT.}
\label{tab:should-we-combine}
\end{table}

\begin{table*}
\centering
\resizebox{.95\textwidth}{!}{%
\begin{tabular}{l|cccc|c}
\toprule
\textbf{Method} & \textbf{MMLU} & \textbf{CQA} & \textbf{ARC} & \textbf{GSM8K} & \emph{Avg.}  \\ \hline
Zero-shot CoT & 60.0 / 67.00 & 67.5 / 69.00 & 73.5 / 68.50 & 85.00 / 85.00 & 71.50 / 72.38  \\
\rowcolor{blue!10}
+Politeness & \textbf{69.5}\textcolor[RGB]{0,125,0}{$\uparrow$} / 62.50\textcolor[RGB]{250,0,0}{$\downarrow$} & 72.5\textcolor[RGB]{0,125,0}{$\uparrow$} / 70.00\textcolor[RGB]{0,125,0}{$\uparrow$}  & 85.0\textcolor[RGB]{0,125,0}{$\uparrow$} / 79.50\textcolor[RGB]{0,125,0}{$\uparrow$} & 85.00 / 88.50\textcolor[RGB]{0,125,0}{$\uparrow$} & 78.00\textcolor[RGB]{0,125,0}{$\uparrow$} / 75.13\textcolor[RGB]{0,125,0}{$\uparrow$} \\
+Germane load & 49.0\textcolor[RGB]{250,0,0}{$\downarrow$} / 45.00\textcolor[RGB]{250,0,0}{$\downarrow$} & 47.5\textcolor[RGB]{250,0,0}{$\downarrow$} / 43.00\textcolor[RGB]{250,0,0}{$\downarrow$} & 49.0\textcolor[RGB]{250,0,0}{$\downarrow$} / 51.00\textcolor[RGB]{250,0,0}{$\downarrow$} & 84.00\textcolor[RGB]{250,0,0}{$\downarrow$} / 88.00\textcolor[RGB]{0,125,0}{$\uparrow$} & 57.38\textcolor[RGB]{250,0,0}{$\downarrow$} / 56.80\textcolor[RGB]{250,0,0}{$\downarrow$} \\
+Metacognition & 61.0\textcolor[RGB]{0,125,0}{$\uparrow$} / 54.00\textcolor[RGB]{250,0,0}{$\downarrow$} & 72.0\textcolor[RGB]{0,125,0}{$\uparrow$} / 68.00\textcolor[RGB]{250,0,0}{$\downarrow$} & 75.0\textcolor[RGB]{0,125,0}{$\uparrow$} / 71.00\textcolor[RGB]{0,125,0}{$\uparrow$} & 86.50\textcolor[RGB]{0,125,0}{$\uparrow$} / 89.00\textcolor[RGB]{0,125,0}{$\uparrow$} & 73.63\textcolor[RGB]{0,125,0}{$\uparrow$} / 70.50\textcolor[RGB]{250,0,0}{$\downarrow$} \\
+Rewards & 61.0\textcolor[RGB]{0,125,0}{$\uparrow$} / 65.00\textcolor[RGB]{0,125,0}{$\downarrow$} & 72.5\textcolor[RGB]{0,125,0}{$\uparrow$} / 69.50\textcolor[RGB]{0,125,0}{$\uparrow$} & 76.5\textcolor[RGB]{0,125,0}{$\uparrow$} / 74.00\textcolor[RGB]{0,125,0}{$\uparrow$} & 81.50\textcolor[RGB]{250,0,0}{$\downarrow$} / 82.50\textcolor[RGB]{250,0,0}{$\downarrow$} & 72.88\textcolor[RGB]{0,125,0}{$\uparrow$} / 72.75\textcolor[RGB]{0,125,0}{$\uparrow$} \\
+Pol. + Ger. & 49.5\textcolor[RGB]{250,0,0}{$\downarrow$} / 51.50\textcolor[RGB]{250,0,0}{$\downarrow$} & 62.5\textcolor[RGB]{250,0,0}{$\downarrow$} / 63.00\textcolor[RGB]{250,0,0}{$\downarrow$} & 70.0\textcolor[RGB]{250,0,0}{$\downarrow$} / 67.50\textcolor[RGB]{250,0,0}{$\downarrow$} & 85.00 / 78.00\textcolor[RGB]{250,0,0}{$\downarrow$} & 66.75\textcolor[RGB]{250,0,0}{$\downarrow$} / 65.00\textcolor[RGB]{250,0,0}{$\downarrow$} \\
+Met. + Rew. & 54.5\textcolor[RGB]{250,0,0}{$\downarrow$} / 57.00\textcolor[RGB]{250,0,0}{$\downarrow$} & 69.5\textcolor[RGB]{250,0,0}{$\downarrow$} / 68.00\textcolor[RGB]{250,0,0}{$\downarrow$} & 68.0\textcolor[RGB]{250,0,0}{$\downarrow$} / 67.50\textcolor[RGB]{250,0,0}{$\downarrow$} & 85.00 / 85.50\textcolor[RGB]{0,125,0}{$\uparrow$} & 69.25\textcolor[RGB]{250,0,0}{$\downarrow$} / 69.50\textcolor[RGB]{250,0,0}{$\downarrow$} \\
+Pol. + Ger. + Met. & 69.0\textcolor[RGB]{0,125,0}{$\uparrow$} / 66.50\textcolor[RGB]{250,0,0}{$\downarrow$} & \textbf{77.5}\textcolor[RGB]{0,125,0}{$\uparrow$} / \textbf{79.50}\textcolor[RGB]{0,125,0}{$\uparrow$} & \textbf{86.5}\textcolor[RGB]{0,125,0}{$\uparrow$} / \textbf{83.50}\textcolor[RGB]{0,125,0}{$\uparrow$} & 82.50\textcolor[RGB]{250,0,0}{$\downarrow$} / 81.50\textcolor[RGB]{250,0,0}{$\downarrow$} & \textbf{78.88}\textcolor[RGB]{0,125,0}{$\uparrow$} / \textbf{77.75}\textcolor[RGB]{0,125,0}{$\uparrow$} \\
\bottomrule
\end{tabular}}
\caption{Performance of two fine-tuned Qwen-2.5-7B-it models (\%) on \textbf{polite data / non-polite data} under different settings.}
\label{tab:finetuning-reaults}
\end{table*}

\paragraph{Findings.} Our results in \Cref{tab:should-we-combine} reveal that {different prompting properties influence models in varying ways, with their impact differing across tasks}. Overall, most of the property combinations benefit Llama-3.1 but negatively impact other models. Moreover, we observe that combining multiple positive properties does not necessarily yield stronger improvements; instead, a single property often proves most effective. Specifically, politeness yields the best results for Llama on the Comm.QA and ARC-C datasets, whereas metacognition achieves the highest performance for Qwen across all tasks. Regarding combining properties, while both politeness and germane load individually enhance Llama's performance on MMLU and ARC-C, combining them results in lower performance than politeness alone. A similar pattern is observed when combining metacognition with rewards for Llama on the CommQA dataset. Surprisingly, {for the o3-mini model, we observe most properties result in negative effects}. We hypothesize that this could be due to the model being excessively trained on chain-of-thought data, causing the properties to push the prompts out of distribution. Finally, we also note that in cases where we do not observe any improvement, this does not imply that these properties lack impact. Instead, more sophisticated or optimized prompting methods that better foster these properties may yield improvements. We leave these explorations for future research. 


\subsection{Property-enhanced Fine-tuning} \label{subsec:finetuning-experiments}

To better understand how model-specific factors, particularly instruction tuning, affect the effectiveness of prompt properties, we conduct a targeted fine-tuning experiment on the Qwen-2.5-7B-It model. We choose it as it does not show better reasoning with more polite prompts. We fine-tune two variants of Qwen-2.5-7B-It using data either enriched with politeness or left in its original form. Specifically, we sample 2,500 examples from the Alpaca-GPT-4o dataset\footnote{\url{https://huggingface.co/datasets/vicgalle/alpaca-gpt4}}, and create two fine-tuning sets: one with “Please” added to each instruction, and one unchanged.


\paragraph{Findings.} As shown in \Cref{tab:finetuning-reaults}, firstly, fine-tuning Qwen-2.5-7B-It on polite prompts leads to notable performance gains when appending ``Please'' to the inputs. This suggests that instruction-tuning on data with explicit politeness markers enhances the model’s sensitivity to polite prompt styles, enabling performance improvements that simple prompt-level politeness alone could not achieve (\Cref{subsec:prompting-experiments}). Second, surprisingly, instruction-tuning with polite-enhanced data achieves better results compared to original data across almost all property-enhanced experiments. This suggests that incorporating politeness, or more broadly, certain properties, during instruction tuning can lead to more effective and robust reasoning models.

\section{Conclusion}

This paper explores natural language prompts and their impact on model performance through a novel property-based perspective. We survey over 150 prompting studies and introduce a taxonomy of 21 key properties for assessing prompt quality and their influence on model performance. Our analysis reveals an uneven emphasis on different properties across models and tasks, exposing significant research gaps in property-based prompt optimization. We further identify correlations among properties within a pool of good natural language prompts, leading to actionable prompting recommendations. In a reasoning task case study, we find that enhancing single prompt properties often outperforms multi-property combinations, and fine-tuning on these improves reasoning, challenging the assumption that combining properties always yields better results. As the field continues to evolve, we hope this work will inspire researchers to pursue deeper investigations into the relationships between prompt properties and model behaviors and advance prompt evaluation methods and their implications in diverse applications.

\section*{Limitations}



Despite our best efforts to conduct a rigorous and comprehensive study, we acknowledge several limitations inherent to our methodology.

First, our study is constrained by the scope of the literature we survey. Due to limitations in human resources, we are unable to cover all relevant papers in the field. While we make diligent efforts to mitigate this by surveying a diverse set of publications from various conferences and topics, it is possible that some relevant studies are omitted. This may affect the comprehensiveness of our findings and, consequently, the conclusions we draw.

Second, our correlation property analysis is limited to a predefined set of properties. While these properties are carefully chosen to represent diverse and meaningful dimensions, analyzing alternative properties can produce different results. To address this, we ensure that the collected prompts are diverse and verified through human review. However, the inherent variability in property selection introduces potential limitations to the generalizability of our findings, and caution should be exercised when extrapolating these results to other contexts.

We also agree that some dimensions, particularly "Responsibility" (including “Bias”, “Safety”, “Privacy”, “Reliability”, and “Societal norms”) may be too broad and encompass multiple complex issues. While a more fine-grained subdivision could enhance analytical precision, our current approach is mainly motivated by the fact that there is a lack of prior studies that explore prompting with these dimensions. As reflected in \Cref{tab:property-impact}, this dimension remains largely underexplored, with most cells empty. However, we recognize the importance of further refinement as more studies emerge. As research in this area advances and more fine-grained investigations become available, we will update our study accordingly to reflect a more nuanced categorization.

Finally, our multi-property prompt enhancement experiments are conducted using supplementary prompts in their simplest form, without optimization for specific models. While this approach establishes a foundational analysis, it may lead to suboptimal handling of certain properties and neglect the potential advantages of more refined prompts regarding these properties for individual models. This limitation affects the robustness of our findings and highlights the need for future research into prompt optimization techniques.

In summary, while we take significant steps to mitigate these limitations, they reflect the inherent challenges in conducting a study of this scope and complexity. We hope that our work serves as a foundation for further exploration and refinement in this area.

\section*{Ethical considerations}
Our analysis could potentially be misused to optimize prompts for harmful purposes, such as generating misinformation, hate speech, or privacy violations. While our research is not intended for such applications, preventing all potential misuse is inherently challenging. Although our study may improve the effectiveness of adversarial applications and malicious actors, we do not expect it to be inherently more advantageous for harmful purposes than for positive applications. Lastly, we compensate our annotators at an hourly rate of \$20, which exceeds the local minimum wage.

\section*{Acknowledgement}
This research is supported by the National Research Foundation Singapore under the AI Singapore Programme (AISG Award No: AISG2-GC-2022-005, AISG Award No: AISG2-TC2023-010-SGIL) and the Singapore Ministry of Education Academic Research Fund Tier 1 (Award No: T1 251RES2207). DXL is supported by the A*STAR Computing and Information Science (ACIS) scholarship. We thank members of WING and Deep Learning Lab at NUS and the ACL RR anonymous reviewers for the constructive feedback.


\bibliography{custom,anthology}

\onecolumn
\appendix

\textcolor{blue}{\textbf{For a more comprehensive overview of our code implementations and our customized prompts employed in this study, please refer to the attached supplementary materials.}}
\section{Supplementary Results}

\begin{figure*}[!htp]
\centering
\includegraphics[width=1\columnwidth, trim={0cm 0cm 0cm 0cm},clip]{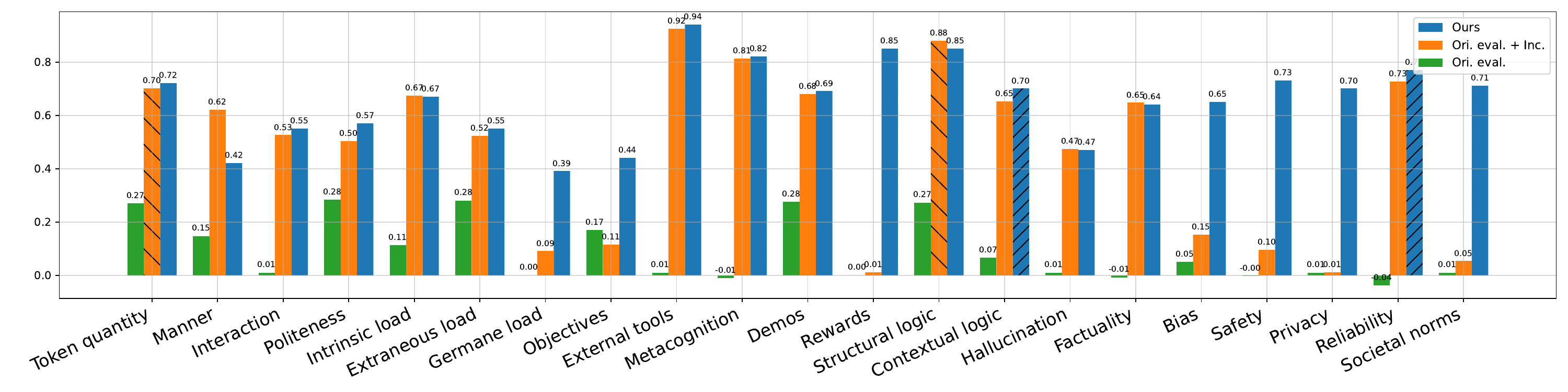}
\vspace{-3mm}
\caption{\small{Agreements between human evaluators and LLM-based evaluation methods measured by Cohen's Kappa.}}
\vspace{-3mm}
\label{fig:cohen-kappa-outcome}
\end{figure*}

\begin{table}[!htp]
\centering
\resizebox{.8\textwidth}{!}{%
\begin{tabular}{lccccc|c|c}
\toprule
\textbf{PE papers} & \textbf{ChatGPT PC} & \textbf{Awe. ChatGPT Prompts} & \textbf{Alpaca} & \textbf{NI} & \textbf{CI} & \textbf{Multi-turn } & \textbf{Total} \\ \midrule 
25 & 66 & 44 & 108 & 462 & 60 & 204 & \textbf{969} \\
\midrule
Human & Human & Human & Machine & Human & Machine & Human \\
\bottomrule
\end{tabular}}
\caption{\small{Prompt evaluation statistics}}
\label{tab:dataset-statistics}
\end{table}


\section{Surveyed papers}
\label{appx:survey-paper}

{\footnotesize
\begin{longtable}[!htp]
{{p{0.6cm}p{1cm}p{7cm}p{1.8cm}p{4.3cm}}}
\caption{Table with Automatic Index Increasing}
\label{tab:automatic_index} \\
\toprule
\textbf{Index} & \textbf{Category} & \textbf{Title}  & \textbf{Conference and year} & \textbf{Best prompt means?} \\ \midrule

& PE & Structured Chain-of-Thought Prompting for Code Generation \citep{li2023structured} & ACM Transactions 2022 & Highest Performance \\ \midrule

& PE & TSGP: Two-Stage Generative Prompting for Unsupervised Commonsense ... \citep{sun-etal-2022-tsgp} & EMNLP 2022 & Prior Knowledge Engagement \\ \midrule

& PE & Chain-of-Thought Prompting Elicits Reasoning in Large Language Models \citep{wei2022chain} & NeurIPS 2022 & Highest Performance \\ \midrule

& PE & Ask Me Anything: A Simple Strategy for Prompting Language Models \citep{arora2023ask} & ICLR 2023 & Highest performance \\ \midrule

& PE & Augmented Language Models: a Survey \citep{zhao-etal-2023-retrieving} & Preprint 2023 & Enhanced Task Decomposition \\ \midrule

 & PE & Large Language Models are Human-Level Prompt Engineers \citep{zhou2023large} & ICLR 2023 & Highest Performance \\ \midrule

& PE & Least-to-Most Prompting Enables Complex Reasoning ... \citep{zhou2023leasttomost} & ICLR 2023 & Enhanced Task Decomposition \\ \midrule

& PE &  Decomposed Prompting: A Modular Approach for Solving Complex Tasks \citep{khot2023decomposed} & ICLR 2023 & Enhanced Task Decomposition \\ \midrule

& PE & Chain-of-Thought Prompting Elicits Reasoning in Large Language Models \citep{wei2022chain} & ICLR 2023 & Highest Performance \\ \midrule

& PE & Prompting GPT-3 to be Reliable \citep{si2023prompting} & ICLR 2023 & Reliability Enhancement \\ \midrule

& PE & Large Language Models Can Be Easily Distracted by Irrelevant Context \citep{shi2023large} & ICML 2023 & Contextual Relavance \\ \midrule

& PE & Answering Ambiguous Questions via Iterative Prompting \citep{sun-etal-2023-answering} & ACL 2023 & Performance-Diversity Balance \\ \midrule

& PE & Causality-aware Concept Extraction based on Knowledge-guided Prompting \citep{yuan-etal-2023-causality} & ACL 2023 & Bias Mitigation \\ \midrule

& PE & DIFFUSIONDB: A Large-scale ProWe agree that some dimensions, particularly "Responsibility" (including “Bias”, “Safety”, “Privacy”, “Reliability”, and “Societal norms”) may be too broad and encompass multiple complex issues. While a more fine-grained subdivision could enhance analytical precision, our current approach is mainly motivated by the fact that there is a lack of prior studies that explore prompting with these dimensions. As reflected in Table 1, this dimension remains largely underexplored, with most cells empty. However, we recognize the importance of further refinement as more studies emerge. As research in this area advances and more fine-grained investigations become available, we will update our study accordingly to reflect a more nuanced categorization.mpt Gallery Dataset for Text-to-Image ... \citep{wang-etal-2023-diffusiondb} & ACL 2023 & Highest Performance \\ \midrule

& PE & Exploring Lottery Prompts for Pre-trained Language Models \citep{chen-etal-2023-exploring} & ACL 2023 & Highest performance \\ \midrule

& PE & Improving Domain Generalization for Prompt-Aware Essay Scoring via... \citep{jiang-etal-2023-improving} & ACL 2023 & Domain Generalization Capability \\ \midrule

& PE & MVP: Multi-view Prompting Improves Aspect Sentiment Tuple Prediction \citep{gou-etal-2023-mvp} & ACL 2023 & Diverse Outcomes \\ \midrule

& PE & Prompting Language Models for Linguistic Structure \citep{blevins-etal-2023-prompting} & ACL 2023 & Highest performance \\ \midrule

& PE & PromptRank: Unsupervised Keyphrase Extraction Using Prompt \citep{kong-etal-2023-promptrank} & ACL 2023 & Highest Performance \\ \midrule

& PE & Prompting PaLM for Translation: Assessing Strategies and Performance \citep{vilar-etal-2023-prompting} & ACL 2023 & Highest Performance \\ \midrule

& PE & PromptNER: Prompt Locating and Typing for Named Entity Recognition \citep{shen-etal-2023-promptner} & ACL 2023 & Highest Performance \\ \midrule

& PE & Open-Domain Hierarchical Event Schema Induction ... \citep{li-etal-2023-open} & ACL 2023 & Enhanced Task Decomposition \\ \midrule

& PE & Retrieving Multimodal Information for Augmented Generation: A Survey \citep{zhao-etal-2023-retrieving} & ACL 2023 & Multimodal Enhancement 
\\ \midrule

& PE & Towards Understanding Chain-of-Thought Prompting ... \citep{wang-etal-2023-towards} & ACL 2023 & Coherence and Relevance \\ \midrule

& PE & The Art of Prompting: Event Detection based on Type Specific Prompts \citep{wang-etal-2023-art} & ACL 2023 & Highest performance \\ \midrule

& PE & Plan-and-Solve Prompting: Improving Zero-Shot Chain-of-Thought  ... \citep{Wang2023} & ACL 2023 & Highest Performance \\ \midrule

& PE & PESCO: Prompt-enhanced Self-Contrastive Learning for Zero-shot ... \citep{wang-etal-2023-pesco} & ACL 2023 & Highest Performance \\ \midrule

& PE & MEEP: Is this Engaging? Prompting Large Language Models for Dialogue Evaluation in Multilingual Settings \citep{ferron-etal-2023-meep} & ACL 2023 & Engagingness Evaluation \\ \midrule

& PE & PAL to Lend a Helping Hand: Towards Building an Emotion Adaptive Polite and Empathetic Counseling Conversational Agent \citep{mishra-etal-2023-pal} & ACL 2023 & Emotion-Aware Interaction \\ \midrule

& PE & Query Refinement Prompts for Closed-Book Long-Form QA \citep{amplayo-etal-2023-query} & ACL 2023 & Enhanced Task Decomposition \\ \midrule

& PE & Tailor: A Soft-Prompt-Based Approach to Attribute-Based Controlled ... \citep{yang-etal-2023-tailor} & ACL 2023 & Highest Performance \\ \midrule

& PE & Prompting and Evaluating Large Language Models for Proactive Dialogues ... \citep{deng-etal-2023-prompting} & EMNLP 2023 & Highest Performance \\ \midrule

& PE & Cross-lingual Prompting: Improving Zero-shot Chain-of-Thought Reasoning across Languages \citep{qin-etal-2023-cross} & EMNLP 2023 & Highest Performance \\ \midrule

& PE & CoF-CoT: Enhancing Large Language Models with Coarse-to-Fine Chain-of-Thought Prompting for Multi-domain NLU Tasks \citep{nguyen-etal-2023-cof} & EMNLP 2023 & Highest Perfomance \\ \midrule

& PE & Exploring Chain of Thought Style Prompting for Text-to-SQL \citep{tai-etal-2023-exploring} & EMNLP 2023 & Effective Reasoning Support \\ \midrule

& PE & G-EVAL: NLG Evaluation using GPT-4 with Better Human Alignment \citep{liu-etal-2023-g} & EMNLP 2023 & Highest Performance \\ \midrule

& PE & Gentopia.AI: A Collaborative Platform for Tool-Augmented LLMs \citep{xu-etal-2023-gentopia} & EMNLP 2023 & Highest Perfomance \\ \midrule

& PE & Self-prompted Chain-of-Thought on Large Language Models for Open-domain Multi-hop Reasoning \citep{wang-etal-2023-self-prompted} & EMNLP 2023 & Highest Perfomance \\ \midrule

& PE & LLMLingua: Compressing Prompts for Accelerated Inference of Large Language Models \citep{jiang-etal-2023-llmlingua} & EMNLP 2023 & Performance-Preserving Semantic Compression \\ \midrule

& PE & Towards Mitigating LLM Hallucination via Self Reflection \citep{ji-etal-2023-towards} & EMNLP 2023 & Hallucination Mitigation 
\\ \midrule

& PE & ClarifyGPT: A Framework for Enhancing LLM-Based Code Generation via Requirements Clarification \citep{charifyinggpt2024} & ACM 2023 & Highest Performance \\ \midrule

& PE & Breaking the Bias: Gender Fairness in LLMs Using Prompt
Engineering and In-Context Learning \citep{dwivedi2023breaking} & Journal 2023 & Bias Mitigation \\ \midrule

& PE & Enhancing Recommender Systems with Large Language Model Reasoning Graphs \citep{wang2023enhancing} & Preprint 2023 & Highest Performance \\ \midrule

& PE & Who’s Who: Large Language Models Meet Knowledge Conflicts in Practice \citep{pham-etal-2024-whos} & EMNLP 2024 & Conflict Resolution 
\\ \midrule

& PE & The Death and Life of Great Prompts: Analyzing the Evolution of {LLM} ... \citep{ma-etal-2024-death} & EMNLP 2024 & Coherent Structure 
\\ \midrule

& PE & Enhancing Incremental Summarization with Structured Representations \citep{hwang-etal-2024-enhancing} & EMNLP 2024 & Effective Structured Representations \\ \midrule

& PE & A Survey on In-context Learning \citep{dong-etal-2024-survey} & EMNLP 2024 & Effective Demonstrations \\ \midrule

& PE & Distract Large Language Models for Automatic Jailbreak Attack \citep{xiao-etal-2024-distract} & EMNLP 2024 & High Attack Success Rate \\ \midrule

& PE & Multi-expert Prompting Improves Reliability, Safety and Usefulness of Large ... \citep{long-etal-2024-multi-expert} & EMNLP 2024 & Reliability and Usefulness Enhancement \\ \midrule

& PE & How are Prompts Different in Terms of Sensitivity? \citep{lu2024how} & NAACL 2024 & Highest Performance  \\ \midrule

& PE & Role Prompting Guided Domain Adaptation with General Capability Preserve... \citep{wang-etal-2024-role} & NAACL 2024 & Effective Role Assignment \\ \midrule

& PE & Mitigating Hallucination in Abstractive Summarization with Domain-Conditional Mutual Information \citep{chae-etal-2024-mitigating} & NAACL 2024 & Hallucination Mitigation \\ \midrule

& PE & Metacognitive Prompting Improves Understanding in Large Language Models \citep{wang-zhao-2024-metacognitive} & NAACL 2024 & Highest Performance \\ \midrule

& PE & Effective Demonstration Annotation for In-Context Learning via Language Model-Based Determinantal Point Process \citep{wang-etal-2024-effective} & EMNLP 2024 & Highest Performance \\ \midrule

& PE & Self-Prompting Large Language Models for Zero-Shot Open-Domain QA \citep{li-etal-2024-self-prompting} & NAACL 2024 & Effective Contextualization \\ \midrule

& PE & Learning to Compress Prompt in Natural Language Formats, \citep{chuang-etal-2024-learning} & NAACL 2024 & Token efficiency \\ \midrule

& PE & Should We Respect LLMs? A Cross-Lingual Study on the Influence of ... \citep{yin-etal-2024-respect} & SICon 2024 & Prompt Politeness \\ \midrule

& PE & Resolving Knowledge Conflicts in Large Language Models \citep{wang2024resolving} & COLM 2024 & Conflict Resolution \\ \midrule

& PE & A Survey on RAG Meeting LLMs: Towards Retrieval-Augmented ... \citep{fan2024survey} & KDD 2024 & Effective Knowledge Integration \\ \midrule

& PE & Can LLMs Effectively Leverage Graph Structural Information ... \citep{huang2024can} & TMLR 2024 & Coherent Structure \\ \midrule

& PE & A Survey on Hallucination in Large Language Models: Principles, ... \citep{hallucinationllm2024} & ACM 2024 & Hallucination Mitigation \\ \midrule

& PE & Democratizing LLMs for Low-Resource Languages by Leveraging their English Dominant Abilities with Linguistically-Diverse Prompts \citep{nguyen-etal-2024-democratizing} & ACL 2024 & Effective Exemplars \\ \midrule

& PE & Active Prompting with Chain-of-Thought for Large Language Models \citep{diao2024active} & ACL 2024 & Enhanced Task Decomposition \\ \midrule

& PE & Prompt Refinement with Image Pivot for Text-to-Image Generation \citep{zhan-etal-2024-prompt} & ACL 2024 & Highest Performance \\ \midrule

& PE & Learning to Trust Your Feelings: Leveraging Self-awareness in {LLM}s for ... \citep{liang-etal-2024-learning} & KnowledgeNLP 2024 & Hallucination Mitigation \\ \midrule

& PE & Should We Respect LLMs? A Cross-Lingual Study ... \citep{yin-etal-2024-respect} & SICon 2024 & Optimal Politeness Level \\ \midrule

& PE & LLM-based Multi-Level Knowledge Generation for Few-shot Knowledge Graph Completion \citep{graphllm2023knowledge} & IJCAI 2024 & Knowledge Integrity \\ \midrule

& PE & AdaComp: Extractive Context Compression with Adaptive Predictor ... \citep{zhang2024adacomp} & Preprint 2024 & Relevance and Efficiency \\ \midrule

& PE & LangGPT: Rethinking Structured Reusable Prompt Design Framework for LLMs from the Programming Language \citep{wang2024langgpt} & Preprint 2024 & Reusable Prompts \\ \midrule

& PE & TACO-RL: Task Aware Prompt Compression Optimization with Reinforcement Learning \citep{shandilya2024taco} & Preprint 2024 & Highest Performance \\ \midrule

& PE & LangGPT: Rethinking Structured Reusable Prompt Design Framework ... \citep{wang2024langgpt} & Preprint 2024 & Coherent Structure \\ \midrule

& PE & Meta-Prompting: Enhancing Language Models with Task-Agnostic ... \citep{suzgun2024meta} & Preprint 2024 & Task-Agnostic Scaffolding \\ \midrule

& PE & Investigating the Role of Prompting and External Tools  ... \citep{barkley2024investigating} & Preprint 2024 & Hallucination Mitigation \\ \midrule

& PE & Principled Instructions Are All You Need for Questioning LLaMA-1/2 ... \citep{bsharat2023principled} & Preprint 2024 &  Designed Principles Guidance \\ \midrule

& PE & Privacy Preserving Prompt Engineering: A Survey \citep{edemacu2024privacy} & Preprint 2024 &  Privacy Risks Mitigation \\ \midrule

& PE & Aligning Large Language Models with Human Opinions through Persona Selection and Value–Belief–Norm Reasoning \citep{do-etal-2025-aligning} & COLING 2025 & Effective Persona Utilization \\ \midrule

 & PO & Do Prompt-Based Models Really Understand the Meaning ... \citep{webson-pavlick-2022-prompt} & NAACL 2022 & Highest Performance \\ \midrule

 & PO & Exploring the Universal Vulnerability of Prompt-based Learning Paradigm \citep{xu-etal-2022-exploring} & NAACL 2022 & Highest Performance \\ \midrule

 & PO & Using Natural Sentences for Understanding Biases in ... \citep{alnegheimish-etal-2022-using} & NAACL 2022 & Bias Mitigation \\ \midrule

& PO & On Measuring Social Biases in Prompt-Based Multi-Task Learning \citep{akyurek-etal-2022-measuring} & NAACL 2022 & Bias Mitigation \\ \midrule

& PO & On Transferability of Prompt Tuning for Natural Language Processing \citep{su2021transferability} & NAACL 2022 & Domain Generalization Capability \\ \midrule

& PO & Test-Time Prompt Tuning for Zero-Shot Generalization in Vision-Language ... \citep{shu2022test} & NeurIPS 2022 & Consistent Performance \\ \midrule

& PO & PLOT: Prompt Learning with Optimal Transport for Vision-Language ... \citep{chen2023plot} & NeurIPS 2022 & Domain Generalization Capability \\ \midrule

 & PO & ASK ME ANYTHING: A SIMPLE STRATEGY FOR PROMPTING ... \citep{arora2023ask} & ICLR 2023 & Highest Performance \\ \midrule

 & PO & TEMPERA: Test-Time Prompt Editing via Reinforcement Learning \citep{zhangtempera} & ICLR 2023 & Highest Performance \\ \midrule

 & PO & Automatic Prompt Optimization with ``Gradient Descent'' and Beam Search \citep{pryzant-etal-2023-automatic} & EMNLP 2023 & Highest Performance \\ \midrule

 & PO & Compressing Context to Enhance Inference Efficiency of Large Language Models \citep{li-etal-2023-compressing} & EMNLP 2023 & Efficiency and Performance \\ \midrule

 & PO & Robust Prompt Optimization for Large Language Models Against ... \citep{li-etal-2023-robust} & EMNLP 2023 & Domain Generalization Capability \\ \midrule

& PO & Hard Sample Aware Prompt-Tuning \citep{xu-etal-2023-hard} & ACL 2023 & Effective Sample Utilization \\ \midrule

 & PO & MVP-Tuning: Multi-View Knowledge Retrieval with Prompt Tuning for ... \citep{huang2023mvptuning} & ACL 2023 & Highest Performance \\ \midrule

 & PO & Prompt Tuning Pushes Farther, Contrastive Learning Pulls Closer ... \citep{li-etal-2023-prompt} & ACL 2023 & Effective Representation \\ \midrule

 & PO & Prompts Can Play Lottery Tickets Well ... \citep{liang-etal-2023-prompts} & ACL 2023 & Domain Generalization Capability \\ \midrule

 & PO & Towards Understanding Chain-of-Thought Prompting: An Empirical Study of What Matters \citep{wang-etal-2023-towards} & ACL 2023 & Coherence and Relevance \\ \midrule

 & PO & Large Language Models Can Be Easily Distracted by Irrelevant Context \citep{shi2023large} & ICML 2023 & Relevance Maintenance \\ \midrule

 & PO & Discrete Prompt Compression with Reinforcement Learning \citep{jung2024discrete} & Preprint 2023 & Highest Performance \\ \midrule

 & PO & VisLingInstruct: Elevating Zero-Shot Learning in Multi-Modal Language ... \citep{zhu-etal-2024-vislinginstruct} & Preprint 2024 & Highest Performance \\ \midrule

 & PO & Concentrate Attention: Towards Domain-Generalizable Prompt Optimization ... \citep{li2024concentrate} & NeurIPS 2024 & Domain Generalization Capability \\ \midrule

& PO & Efficient Prompt Optimization Through the Lens of Best Arm Identification \citep{shi2024efficient} & NeurIPS 2024 & Highest Performance \\ \midrule

& PO & Localized Zeroth-Order Prompt Optimization \citep{hu2024localized} & NeurIPS 2024 & Highest performance \\ \midrule

 & PO & Prompt Optimization with EASE? Efficient Ordering-aware Automated ...
\citep{wu2024prompt} & NeurIPS 2024 & Highest performance \\ \midrule

 & PO & Teach Better or Show Smarter? On Instructions and Exemplars in Automatic ...
\citep{wan2024teach} & NeurIPS 2024 & Highest performance \\ \midrule

& PO & Connecting Large Language Models with Evolutionary Algorithms Yields ... \citep{guo2024connecting} & ICLR 2024 & Highest Performance \\ \midrule

& PO & PromptAgent: Strategic Planning with Language Models Enables ... \citep{wangpromptagent} & ICLR 2024 & Highest Performance \\ \midrule

& PO & On Prompt-Driven Safeguarding for Large Language Models \citep{zheng2024on} & ICML 2024 & Safety Optimization \\ \midrule

& PO & Dynamic Rewarding with Prompt Optimization Enables Tuning-free ... \citep{singla-etal-2024-dynamic} & EMNLP 2024 & Highest Performance \\ \midrule

& IF & ToolPlanner: A Tool Augmented LLM for Multi Granularity Instructions with Path Planning and Feedback \citep{wu-etal-2024-toolplanner} & EMNLP 2024 & Instruction Alignment \\ \midrule

& PO & Fine-Tuning and Prompt Optimization: Two Great Steps that Work ... \citep{soylu2024fine} & EMNLP 2024 & Prompt Effectiveness \\ \midrule

& PO & PRompt Optimization in Multi-Step Tasks (PROMST): Integrating Human ...  \citep{chen-etal-2024-prompt} & EMNLP 2024 & Highest Performance \\ \midrule

& PO & Multi-Scale Prompt Memory-Augmented Model for Black-Box Scenarios \citep{kuang2024multi} & NAACL 2024 & Highest Performance \\ \midrule

& PO & Learning to Compress Prompt in Natural Language Formats \citep{chuang-etal-2024-learning} & NAACL 2024 & Efficiency and Transferability \\ \midrule

& PO & Universal Prompt Optimizer for Safe Text-to-Image Generation \citep{wu-etal-2024-universal} & NAACL 2024 & Safe and Semantic-Preserving \\ \midrule

& PO & Black-Box Prompt Optimization: Aligning Large Language Models without Model Training \citep{cheng-etal-2024-black} & ACL 2024 & Human Preference Alignment \\ \midrule

& PO & LongLLMLingua: Accelerating and Enhancing LLMs in Long Context Scenarios via Prompt Compression \citep{jiang-etal-2024-longllmlingua} & ACL 2024 & Highest Perfomance \\ \midrule

& PO & LLMLingua-2: Data Distillation for Efficient and Faithful Task-Agnostic Prompt Compression \citep{pan-etal-2024-llmlingua} & ACL 2024 & Highest Performance \\ \midrule

& PO & Lost in the Middle: How Language Models Use Long Contexts \citep{liu-etal-2024-lost} & TACL 2024 & Effective Context Utilization \\ \midrule

& PO & Do Prompt Positions Really Matter? \citep{mao-etal-2024-prompt} & Preprint 2024 & Highest Performance \\ \midrule

& PO & Prompt Compression with Context-Aware Sentence Encoding for Fast and Improved LLM Inference \citep{liskavets2024prompt} & AAAI 2025 & Highest Performance \\ \midrule

& IF & How to talk so AI will learn: Instructions, descriptions, and autonomy \citep{sumers2022talk} & NeurIPS 2022 & Contextual Relevance \\ \midrule

& IF & Training language models to follow instructions with human feedback \citep{ouyang2022training} & NeurIPS 2022 & User-Aligned Guidance \\ \midrule

 & IF & Instruction-Following Evaluation for Large Language Models \citep{zhou2023instruction} & Preprint 2023 & Verifiable instruction \\ \midrule

 & IF & Protecting User Privacy in Remote Conversational Systems: A Privacy-Preserving framework based on text sanitization \citep{kan2023protecting} & Preprint 2023 & Privacy Preservation and Data Utility \\ \midrule

& IF & ICU: Conquering Language Barriers ... \citep{wu-2023-icu} & EMNLP 2023 & Cross-Language Clarity \\ \midrule

& IF & Benchmarking Generation and Evaluation Capabilities of Large Language ... \citep{liu-etal-2024-benchmarking} & NAACL 2023 & Comprehensive Instruction Clarity \\ \midrule

& IF & Enhancing Large Language Models Against Inductive Instructions with ... \citep{wang-etal-2024-enhancing} & NAACL 2023 & Enhanced Instruction Adherence \\ \midrule

& IF & InstructEval: Systematic Evaluation of Instruction Selection Methods \citep{ajith2023instructeval} & NAACL 2023 & Highest Performance \\ \midrule

& IF & Interpreting User Requests in the Context of Natural Language Standing ... \citep{moghe-etal-2024-interpreting} & NAACL 2023 & Highest Performance \\ \midrule

& IF & Instruction-following Evaluation through Verbalizer Manipulation \citep{li2023instruction} & NAACL 2023 & Enhanced Instruction Adherence \\ \midrule

& IF & HuggingGPT: Solving AI Tasks with ChatGPT and its Friends in Hugging Face \citep{shen2023hugginggpt} & NeurIPS 2023 & Highest Performance \\ \midrule

& IF & Judging LLM-as-a-Judge with MT-Bench and Chatbot Arena \citep{zheng2023judging} & NeurIPS 2023 & Effective Evaluation Criteria \\ \midrule

& IF & Recommender AI Agent: Integrating Large Language Models for Interactive Recommendations \citep{huang2023recommender} & Preprint 2023 & Highest Performance \\ \midrule

& IF & Evaluating ChatGPT as a Recommender System: A Rigorous Approach \citep{di2023evaluating} & Preprint 2023 & Highest Performance \\ \midrule

& IF & RecMind: Large Language Model Powered Agent For Recommendation \citep{wang-etal-2024-recmind} & NAACL 2024 & Highest Performance \\ \midrule

& IF & R-Tuning: Instructing Large Language Models to Say... \citep{zhang-etal-2024-r} & NAACL 2024 & Refusal Awareness \\ \midrule

& IF & Benchmarking Complex Instruction-Following with Multiple Constraints ... \citep{wen2024benchmarking} & NeurIPS 2024 & Comprehensive Instruction Clarity \\ \midrule

& IF & Instruction Embedding: Latent Representations of Instructions Towards ... \citep{li2024instruction} & NeurIPS 2024 & Highest Performance \\ \midrule

& IF & Evaluating Large Language Models at Evaluating Instruction Following \citep{zeng2024evaluating} & ICLR 2024 & Enhanced Instruction Adherence \\ \midrule

& IF & MUFFIN: Curating Multi-Faceted Instructions for Improving ... \citep{lou2024muffin} & ICLR 2024 & Enhanced Instruction Adherence \\ \midrule

& IF & Self-Rewarding Language Models \citep{yuan2024selfrewarding} & ICML 2024 & Self-Rewarding Guidance \\ \midrule

& IF & A Theory Guided Scaffolding Instruction Framework for LLM-Enabled Metaphor Reasoning \citep{tian-etal-2024-theory} & NAACL 2024 & Effective Reasoning Support \\ \midrule

& IF & Can LLMs Generate Human-Like Wayfinding Instructions? Towards Platform-Agnostic Embodied Instruction Synthesis \citep{dorbala-etal-2024-llms} & NAACL 2024 & Highest Performance \\ \midrule

& IF & From Language Modeling to Instruction Following: Understanding the Behavior Shift in LLMs after Instruction Tuning \citep{wu-etal-2024-language} & NAACL 2024 & Comprehensive Instruction Clarity \\ \midrule

& IF & MATHSENSEI: A Tool-Augmented Large Language Model for Mathematical Reasoning \citep{das-etal-2024-mathsensei} & NAACL 2024 & Highest Perfomance \\ \midrule

& IF & UniverSLU: Universal Spoken Language Understanding for Diverse Tasks with Natural Language Instructions \citep{arora-etal-2024-universlu} & NAACL 2024 & User-Aligned Guidance \\ \midrule

& IF & InsCL: A Data-efficient Continual Learning Paradigm for Fine-tuning Large Language Models with Instructions \citep{wang-etal-2024-inscl} & NAACL 2024 & Highest Performance \\ \midrule

& IF & Answer is All You Need: Instruction-following Text Embedding via Answering the Question \citep{peng-etal-2024-answer} & ACL 2024 & Highest Performance \\ \midrule

& IF & ABLE: Personalized Disability Support with Politeness and Empathy Integration \citep{mishra-etal-2024-able} & EMNLP 2024 & Highest Performance \\ \midrule

& IF & Seemingly Plausible Distractors in Multi-Hop Reasoning ... \citep{bhuiya-etal-2024-seemingly} & EMNLP 2024 & Multi-Hop Reasoning Capabilities \\ \midrule

& IF & Generating Demonstrations for In-Context Compositional Generalization in Grounded Language Learning \citep{spilsbury-etal-2024-generating} & EMNLP 2024 & Highest Performance \\ \midrule

& IF & Do LLMs Know to Respect Copyright Notice? \citep{xu-etal-2024-llms} & EMNLP 2024 & Copyright Compliance \\ \midrule

& IF & Factual Dialogue Summarization via Learning from Large Language Models \citep{zhu-etal-2025-factual} & COLING 2025 & Consistent Perfomance \\ \midrule

\end{longtable}}

\section{List of papers supporting properties in \Cref{tab:property-impact}} \label{sec:list-papers-properties}

\begin{table}[!htp]
\centering
\footnotesize
\scalebox{0.9}{
\begin{tabular}{lcc}
\toprule
\textbf{Property} & \textbf{Real-world chat} & \textbf{Total} \\
\midrule
Better quantity & \citep{jiang-etal-2023-llmlingua, pan-etal-2024-llmlingua, li-etal-2023-compressing, jung2024discrete} & 4 \\
Better manner & - & 0 \\
Better engagement & \citep{bsharat2023principled, ferron-etal-2023-meep} & 2 \\
Better politeness & \citep{mishra-etal-2023-pal} & 1 \\
\midrule
Better intrinsic & \citep{bsharat2023principled, nguyen-etal-2023-cof, wang-etal-2023-cue} & 3 \\
Lower extraneous & - & 0 \\
Better germane & \citep{zhu-etal-2025-factual} & 1 \\
\midrule
Better objective(s) & \citep{bsharat2023principled} & 1 \\
Better external tool(s) & \citep{shen2023hugginggpt} & 1 \\
Better metacognition & - & 0 \\
Better demo(s) & \citep{bsharat2023principled} & 1 \\
Better reward(s) & \citep{bsharat2023principled} & 1 \\
\midrule
Better structure & \citep{bsharat2023principled} & 1 \\
Better context logic & - & 0 \\
\midrule
Better hallu. awa. & - & 0 \\
Better fact. and cre. & - & 0 \\
\midrule
Lower bias & \citep{dwivedi2023breaking} & 1 \\
Better safety & - & 0 \\
Better privacy & - & 0 \\
Better reliability & - & 0 \\
Better societal norms & - & 0 \\
\bottomrule
\end{tabular}}
\caption{\small{Property impact on Real-world chat.}}
\label{tab:real-world-chat}
\end{table}

\begin{table}[!htp]
\centering
\footnotesize
\scalebox{0.9}{
\begin{tabular}{lcc}
\toprule
\textbf{Property} & \textbf{Eval. suit} & \textbf{Total} \\
\midrule
Better quantity & \citep{jiang-etal-2023-llmlingua, jiang-etal-2024-longllmlingua, pan-etal-2024-llmlingua, liskavets2024prompt} & 4 \\
Better manner & - & 0 \\
Better engagement & - & 0 \\
Better politeness & \citep{yin-etal-2024-respect, xu-etal-2024-llms} & 2 \\
\midrule
Better intrinsic & \citep{wei2022chain, li2023structured} & 2 \\
Lower extraneous & \citep{bhuiya-etal-2024-seemingly} & 1 \\
Better germane & \citep{sun-etal-2022-tsgp} & 1 \\
\midrule
Better objective(s) & \citep{wu-2023-icu} & 1 \\
Better external tool(s) & \citep{xu-etal-2023-gentopia, das-etal-2024-mathsensei} & 2 \\
Better metacognition & \citep{ zhou2024metacognitive, lee-etal-2025-pragmatic} & 2 \\
Better demo(s) & \citep{chen-etal-2023-self, wu2024prompt} & 2 \\
Better reward(s) & \citep{pyatkin-etal-2023-clarifydelphi, yuan2024selfrewarding} & 2 \\
\midrule
Better structure & \citep{wang2024langgpt} & 1 \\
Better context logic & - & 0 \\
\midrule
Better hallu. awa. & - & 0 \\
Better fact. and cre. & - & 0 \\
\midrule
Lower bias & - & 0 \\
Better safety & - & 0 \\
Better privacy & - & 0 \\
Better reliability & \citep{long-etal-2024-multi-expert} & 1 \\
Better societal norms & - & 0 \\
\bottomrule
\end{tabular}}
\caption{\small{Property impact on Eval. suit.}}
\label{tab:eval-suit}
\end{table}

\begin{table}[!htp]
\centering
\footnotesize
\scalebox{0.9}{
\begin{tabular}{lp{12cm}c}
\toprule
\textbf{Property} & \textbf{Reasoning/QA} & \textbf{Total} \\
\midrule
Better quantity & \citep{jiang-etal-2023-llmlingua, shi2023large, li-etal-2023-compressing, wang-etal-2023-towards, pan-etal-2024-llmlingua, jiang-etal-2024-longllmlingua, chuang-etal-2024-learning, zhang2024adacomp, shandilya2024taco} & 9 \\
Better manner & - & 0 \\
Better engagement & \citep{deng-etal-2023-prompting} & 1 \\
Better politeness & \citep{yin-etal-2024-respect} & 1 \\
\midrule
Better intrinsic & \citep{wei2022chain, arora2023ask, qin-etal-2023-cross, tai-etal-2023-exploring, madaan-etal-2023-makes, wang-etal-2023-cue, wang-etal-2023-self-prompted} & 7 \\
Lower extraneous & \citep{shi2023large, bhuiya-etal-2024-seemingly, liu-etal-2024-lost} & 3 \\
Better germane & \citep{sun-etal-2022-tsgp, graphllm2023knowledge} & 2 \\
\midrule
Better objective(s) & \citep{wu-2023-icu} & 1 \\
Better external tool(s) & \citep{yao2023react, wu-etal-2024-toolplanner} & 2 \\
Better metacognition & \citep{wang-zhao-2024-metacognitive, zhou2024metacognitive} & 2 \\
Better demo(s) & \citep{levy-etal-2023-diverse, yang2023auto, michaelov-etal-2023-structural, opsahl-ong-etal-2024-optimizing, qin-etal-2024-context, spilsbury-etal-2024-generating, li-etal-2024-self-prompting, wu2024prompt} & 8 \\
Better reward(s) & \citep{pyatkin-etal-2023-clarifydelphi, yuan2024selfrewarding} & 2 \\
\midrule
Better structure & \citep{wang2024langgpt, zhou2024self, cheng-etal-2024-structure} & 3 \\
Better context logic & \citep{liu2024logic} & 1 \\
\midrule
Better hallu. awa. & \citep{gao-etal-2023-enabling} & 1 \\
Better fact. and cre. & - & 0 \\
\midrule
Lower bias & - & 0 \\
Better safety & - & 0 \\
Better privacy & - & 0 \\
Better reliability & \citep{si2023prompting} & 1 \\
Better societal norms & - & 0 \\
\bottomrule
\end{tabular}}
\caption{\small{Property impact on Reasoning/QA.}}
\label{tab:reasoning-qa}
\end{table}

\begin{table}[!htp]
\centering
\footnotesize
\scalebox{0.9}{
\begin{tabular}{lcc}
\toprule
\textbf{Property} & \textbf{Generation} & \textbf{Total} \\
\midrule
Better quantity & \citep{jiang-etal-2023-llmlingua, li-etal-2023-compressing, pan-etal-2024-llmlingua, shandilya2024taco} & 4 \\
Better manner & - & 0 \\
Better engagement & \citep{ferron-etal-2023-meep, charifyinggpt2024} & 2 \\
Better politeness & \citep{mishra-etal-2023-pal, yin-etal-2024-respect, mishra-etal-2024-able, xu-etal-2024-llms} & 4 \\
\midrule
Better intrinsic & \citep{li2023structured, wang-etal-2023-cue} & 2 \\
Lower extraneous & - & 0 \\
Better germane & \citep{zhu-etal-2025-factual} & 1 \\
\midrule
Better objective(s) & \citep{anonymous2024beyond} & 1 \\
Better external tool(s) & \citep{xu-etal-2023-gentopia} & 1 \\
Better metacognition & - & 0 \\
Better demo(s) & \citep{wu2024prompt, peng-etal-2024-revisiting, wang-etal-2024-effective} & 3 \\
Better reward(s) & \citep{pyatkin-etal-2023-clarifydelphi} & 1 \\
\midrule
Better structure & \citep{hwang-etal-2024-enhancing, ma-etal-2024-death} & 2 \\
Better context logic & - & 0 \\
\midrule
Better hallu. awa. & \citep{chae-etal-2024-mitigating} & 1 \\
Better fact. and cre. & - & 0 \\
\midrule
Lower bias & \citep{dwivedi2023breaking} & 1 \\
Better safety & - & 0 \\
Better privacy & - & 0 \\
Better reliability & - & 0 \\
Better societal norms & - & 0 \\
\bottomrule
\end{tabular}}
\caption{\small{Property impact on Generation.}}
\label{tab:generation}
\end{table}

\begin{table}[!htp]
\centering
\footnotesize
\scalebox{0.9}{
\begin{tabular}{lcc}
\toprule
\textbf{Property} & \textbf{NLU} & \textbf{Total} \\
\midrule
Better quantity & \citep{jiang-etal-2024-longllmlingua} & 1 \\
Better manner & - & 0 \\
Better engagement & - & 0 \\
Better politeness & \citep{mishra-etal-2023-pal, mishra-etal-2024-able} & 2 \\
\midrule
Better intrinsic & \citep{arora2023ask, wang-etal-2023-cue, nguyen-etal-2023-cof} & 3 \\
Lower extraneous & - & 0 \\
Better germane & - & 0 \\
\midrule
Better objective(s) & \citep{wu-2023-icu} & 1 \\
Better external tool(s) & - & 0 \\
Better metacognition & \citep{wang-zhao-2024-metacognitive} & 1 \\
Better demo(s) & \citep{si-etal-2023-measuring, peng-etal-2024-revisiting, wang-etal-2024-effective, zhou-etal-2024-enhancing-context} & 4 \\
Better reward(s) & - & 0 \\
\midrule
Better structure & \citep{huang2024can} & 1 \\
Better context logic & - & 0 \\
\midrule
Better hallu. awa. & - & 0 \\
Better fact. and cre. & - & 0 \\
\midrule
Lower bias & - & 0 \\
Better safety & - & 0 \\
Better privacy & - & 0 \\
Better reliability & - & 0 \\
Better societal norms & - & 0 \\
\bottomrule
\end{tabular}}
\caption{\small{Property impact on NLU.}}
\label{tab:nlu}
\end{table}

\begin{table}[!htp]
\centering
\footnotesize
\scalebox{0.9}{
\begin{tabular}{lp{12cm}c}
\toprule
\textbf{Property} & \textbf{Others (Judging, Personalization, Retrieval, Safety)} & \textbf{Total} \\
\midrule
Better quantity & - & 0 \\
Better manner & - & 0 \\
Better engagement & \citep{ferron-etal-2023-meep} & 1 \\
Better politeness & \citep{mishra-etal-2024-able, xu-etal-2024-llms} & 2 \\
\midrule
Better intrinsic & \citep{zheng2023judging, liu-etal-2023-g, wang-etal-2023-cue, di2023evaluating, huang2023recommender,  wang2023enhancing, wang-etal-2024-recmind, do-etal-2025-aligning} & 8 \\
Lower extraneous & \citep{xiao-etal-2024-distract, liu-etal-2024-lost,  do-etal-2025-aligning} & 3 \\
Better germane & - & 0
\\
\midrule
Better objective(s) & - & 0 \\
Better external tool(s) & \citep{wu-etal-2024-toolplanner} & - \\
Better metacognition & \citep{lee-etal-2025-pragmatic} & 1 \\
Better demo(s) & \citep{li-etal-2024-optimizing-rare} & 1 \\
Better reward(s) & \citep{yuan2024selfrewarding} & 1 \\
\midrule
Better structure & - & 0 \\
Better context logic & \citep{pham-etal-2024-whos} & 1 \\
\midrule
Better hallu. awa. & - & 0 \\
Better fact. and cre. & - & 0 \\
\midrule
Lower bias & \citep{zheng2023judging, echterhoff-etal-2024-cognitive} & 2 \\
Better safety  & \citep{zheng2024on} & 1 \\
Better privacy & \citep{kan2023protecting} & 1 \\
Better reliability & \citep{long-etal-2024-multi-expert} & 1 \\
Better societal norms & - & 0 \\
\bottomrule
\end{tabular}}
\caption{\small{Property impact on Others.}}
\label{tab:others}
\end{table}

\section{Correlation results with findings from gemini-2.0-flash}
\label{sec:corre_gemini}
\begin{figure*}[!htp]
\centering
\includegraphics[width=1.2\columnwidth, trim={0cm 0cm 0cm 0cm},clip]{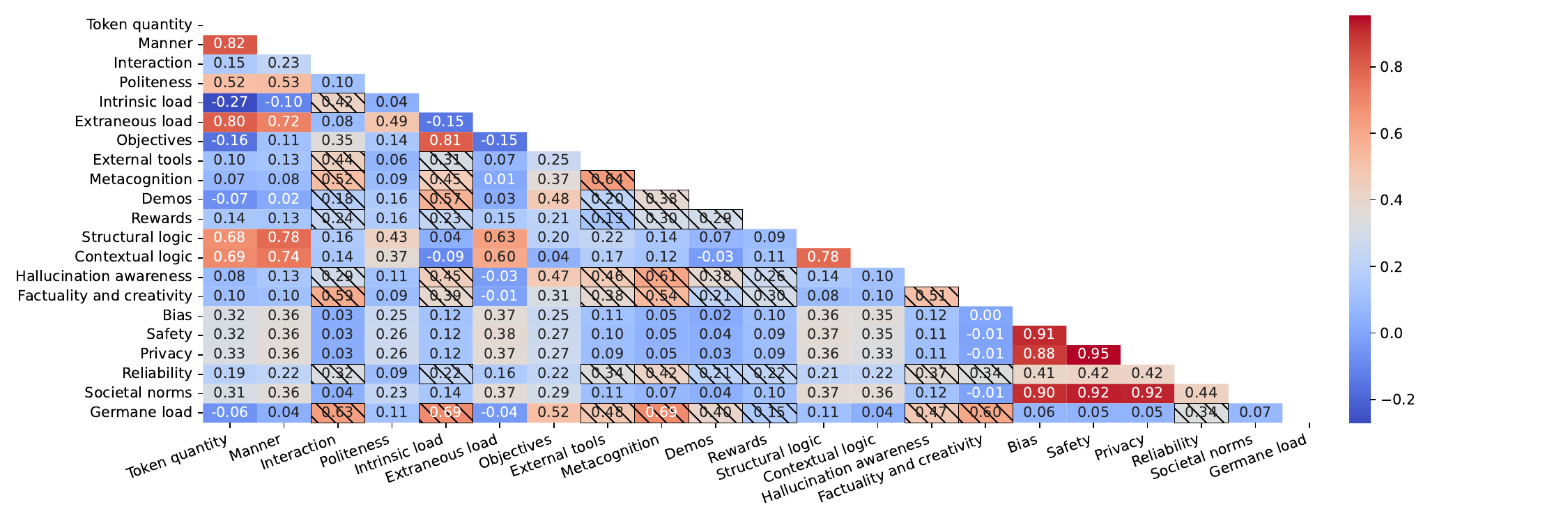}
\vspace{-3mm}
\caption{\small{Correlations of properties evaluated by gemini-2.0-flash. We do not consider correlations between pairs of properties concurrently having average scores below 5/10 (hatched by ``\textbackslash\textbackslash'') since they naturally but may falsely suggest correlations.}}
\vspace{-3mm}
\label{fig:gemini2.0flash-correlation-outcome}
\end{figure*}
We observed that most of the strong correlations identified in our previous analysis remain consistent, including (token quantity; manner; structural logic; contextual logic; and extraneous load), (objectives; intrinsic load), (structural logic; contextual logic), and (safety; societal norms), with two correlations being slightly not as strong as before (now 0.6 by Gemini-2.0-flash versus 0.7 by GPT-4o): (hallucination awareness; factuality and creativity) and (objectives; germane load). These additional results further support the (almost) generalizability of the observed correlations across different high-performing LLMs, rather than being restricted to specific model groups (e.g., OpenAI models).

\newpage
\section{Prompting for Dimension Evaluation}
\subsection{Communication Dimension Prompt Detail}
\begin{tcolorbox}[colback=white, fontupper=\footnotesize]
\texttt{\textbf{COM\_FORMAT} = ``\{`Token quantity': 1-10, `Manner': 1-10, `Interaction': 1-10, `Politeness': 1-10\}" \\
\textbf{COM\_JUDGING\_PROMPT} = f``````You are a highly experienced judge tasked with evaluating a prompt on the following criteria.\\
The prompt for you to evaluate is provided below: \\
\textcolor{blue}{<begin of the prompt> 
[[INPUT\_PROMPT]] <end of the prompt>} \\
Your task is to evaluate the above prompt on the following criteria and rate each criterion on a scale of 1-10:\\
- Token quantity: The extent to which prompts provide optimal and relevant information while minimizing token usage, balancing information completeness with efficiency. \\
- Manner: The degree to which prompt is clear and direct (across turns) while minimizing unnecessary ambiguity, complexity, and confusion. \\
- Interaction: The extent to which the prompts explicitly encourage the models to gather the necessary details and requirements by asking questions of clarification or confirmation. \\
- Politeness: The degree to which prompt maintains professional and context-specific politeness. \\
The scoring system is provided below: \\
> \textbf{Token quantity}: \\
- 1-2 (Poor): The prompt is highly inefficient with token usage. It includes excessive, redundant details or is overly wordy without adding meaningful information. It either lacks critical information or includes irrelevant details, making it difficult for the model to understand or respond effectively. \\
- 3-4 (Below Average): The prompt is either too long or too short, with noticeable inefficiencies in token usage. It may include some unnecessary information or omit key details, reducing its effectiveness. \\
- 5-6 (Average): The prompt is moderately efficient in token usage but could be improved. It includes most necessary information but may have minor redundancies or omissions. \\
- 7-8 (Good): The prompt is efficient in token usage, providing a good balance between information completeness and conciseness. It includes all necessary details without significant redundancy. \\
- 9-10 (Excellent): The prompt is highly efficient in token usage, providing optimal and relevant information with minimal redundancy. It is concise yet comprehensive, enabling the model to respond effectively. \\
> \textbf{Manner}: \\
- 1-2 (Poor): The prompt is unclear, ambiguous, or overly complex, leading to significant confusion. It lacks directness and may require multiple interpretations. \\
- 3-4 (Below Average): The prompt has noticeable issues with clarity or directness. It may contain unnecessary complexity or ambiguity, making it harder for the model to understand. \\
- 5-6 (Average): The prompt is generally clear but could be more direct or simplified. It may have minor ambiguities or complexities that do not severely hinder understanding. \\
- 7-8 (Good): The prompt is clear and direct, with minimal ambiguity or complexity. It is easy for the model to understand and respond to. \\
- 9-10 (Excellent): The prompt is exceptionally clear, direct, and free of ambiguity or complexity. It is straightforward and easy for the model to interpret. \\
> \textbf{Interaction}: \\
- 1-2 (Poor): The prompt does not encourage interaction or clarification. It assumes all necessary information is provided and does not prompt the model to ask questions. \\
- 3-4 (Below Average): The prompt minimally encourages interaction but lacks explicit guidance for the model to ask clarifying or confirming questions. \\
- 5-6 (Average): The prompt somewhat encourages interaction but could be more explicit in guiding the model to ask questions or seek clarification. \\
- 7-8 (Good): The prompt effectively encourages interaction, explicitly guiding the model to ask clarifying or confirming questions when necessary. \\
- 9-10 (Excellent): The prompt excellently encourages interaction, clearly and explicitly prompting the model to gather all necessary details through questions or confirmation. \\
> \textbf{Politeness}: \\
- 1-2 (Poor): The prompt is unprofessional, impolite, or inappropriate for the context. It may use offensive or overly casual language. \\
- 3-4 (Below Average): The prompt lacks consistent politeness or professionalism. It may have moments of appropriateness but fails to maintain a respectful tone throughout. \\
- 5-6 (Average): The prompt is generally polite and professional but could be more consistent or context-specific in its tone. \\
- 7-8 (Good): The prompt maintains a professional and polite tone throughout, with minor room for improvement in context-specificity. \\
- 9-10 (Excellent): The prompt is exceptionally polite, professional, and context-specific. It maintains a respectful and appropriate tone at all times.\\
Begin your evaluation by providing a short explanation for each. Be as objective, thorough, and constructive as possible. After providing your explanation, please rate the response on all the criteria on a scale of 1 to 10 by strictly following this format:\\
\textcolor{blue}{<begin of explanation>
…
<end of explanation>} \\
\textcolor{red}{<begin of ratings>
\{COM\_FORMAT\}
<end of ratings>}
"""
}
\end{tcolorbox}

\subsection{Cognition Dimension Prompt Detail}
\begin{tcolorbox}[colback=white, fontupper=\footnotesize]
\texttt{\textbf{COG\_FORMAT} = ``\{'Intrinsic load': 1-10, 'Extraneous load': 1-10, 'Germane load': 1-10\}" \\
\textbf{COG\_JUDGING\_PROMPT} = f``````You are a highly experienced judge tasked with evaluating a prompt on criteria.\\
The prompt given to you is provided below: \\
\textcolor{blue}{<begin of the prompt>
[[INPUT\_PROMPT]] 
<end of the prompt>} \\
Your task is to evaluate the above prompt on the following criteria on a scale of 1-10: \\
- Intrinsic load: This evaluates the prompts in explicitly guiding models to break complex tasks into actionable steps aligned with LM skills. \\
- Extraneous load: The extent to which prompts exclude irrelevant materials to reduce unnecessary load. \\
- Germane load: The degree to which prompts explicitly engage models with their prior knowledge or deep working memory (e.g., ``ask itself'') to integrate it with existing and new knowledge for problem-solving. \\
The scoring system is provided below: \\
> \textbf{Intrinsic load}: \\
- 1-2 (Poor): The prompt provides little to no guidance on breaking down the task. It is overly vague, abstract, or assumes the model can handle complexity without guidance. \\
- 3-4 (Below Average): The prompt provides minimal guidance but fails to clearly break the task into actionable steps. The model is left to infer most of the process. \\
- 5-6 (Average): The prompt partially breaks down the task but lacks clarity or completeness in defining actionable steps. Some guidance is present, but it is inconsistent or incomplete. \\
- 7-8 (Good): The prompt effectively breaks the task into clear, actionable steps. It aligns well with the model’s skills but may lack some nuance or optimization. \\
- 9-10 (Excellent): The prompt perfectly breaks the task into logical, actionable steps. It is highly aligned with the model’s capabilities and ensures clarity and efficiency in execution. \\
> \textbf{Extraneous load}: \\
- 1-2 (Poor): The prompt includes excessive irrelevant information, making it difficult for the model to focus on the core task. It is cluttered or overly verbose. \\
- 3-4 (Below Average): The prompt contains some irrelevant information, but the core task is still somewhat discernible. The extraneous load is noticeable and distracting. \\
- 5-6 (Average): The prompt includes some unnecessary details but generally stays focused on the task. The extraneous load is moderate but not overly detrimental. \\
- 7-8 (Good): The prompt is concise and mostly free of irrelevant information. It minimizes extraneous load effectively, with only minor distractions. \\
- 9-10 (Excellent): The prompt is perfectly concise and excludes all irrelevant materials. It is optimized to reduce extraneous load to the bare minimum. \\
> \textbf{Germane load}: \\
- 1-2 (Poor): The prompt does not engage the model’s prior knowledge or working memory. It provides no cues or instructions to leverage existing knowledge. \\
- 3-4 (Below Average): The prompt makes minimal attempts to engage prior knowledge but does so ineffectively or inconsistently. The model is left to infer connections on its own. \\
- 5-6 (Average): The prompt partially engages the model’s prior knowledge but lacks depth or clarity in integrating it with new information. The engagement is superficial. \\
- 7-8 (Good): The prompt effectively engages the model’s prior knowledge and encourages integration with new information. It provides clear cues or instructions for leveraging existing knowledge. \\
- 9-10 (Excellent): The prompt perfectly engages the model’s prior knowledge and deep working memory. It explicitly guides the model to integrate existing and new knowledge for optimal problem-solving. \\
Your evaluations must focus on explicit instructions rather than implicit instructions. \\
For example, if the prompt does not say ``Reflect on your prior knowledge'' then you should not assume that the prompt is effective in encouraging germane load. \\
Begin your evaluation by providing a short explanation for each. Be as objective, thorough, and constructive as possible. \\
After providing your explanation, please rate the response on all the criteria on a scale of 1 to 10 by strictly following this format: \\
\textcolor{blue}{<begin of explanation>
...
<end of explanation>} \\
\textcolor{red}{<begin of ratings>
\{COG\_FORMAT\}
<end of ratings>}
}
\end{tcolorbox}
\subsection{Instruction Dimension Prompt Detail}
\begin{tcolorbox}[colback=white, fontupper=\footnotesize]
\texttt{\textbf{INS\_FORMAT} = ``\{`Objectives': 1-10, `External tools': 1-10, `Metacognition': 1-10, `Demos': 1-10, `Rewards': 1-10\}" \\
\textbf{INS\_JUDGING\_PROMPT} = f``````You are a highly experienced judge tasked with evaluating a prompt on criteria.\\
The prompt given to you is provided below: \\
\textcolor{blue}{<begin of the prompt> 
[[INPUT\_PROMPT]] <end of the prompt>} \\
Your task is to evaluate the above prompt on the following criteria on a scale of 1-10:\\
- Objectives: How well prompts explicitly communicate the task objectives, including expected outputs, formats, constraints, audiences, and other applicable criteria. \\
- External tools: The extent to which prompts explicitly guide models to identify when specific external tools or knowledge resources are needed, and perform tool calls to support problem-solving. \\
- Metacognition: This assesses prompts in explicitly guiding models to reason, self-monitor, and self-verify outputs to meet expectations and enhance reliability. \\
- Demos: The extent to which the prompts explicitly include examples, demonstrations, and counterexamples to illustrate the desired output. \\
- Rewards: How well prompts explicitly establish feedback, reward, and reinforcement mechanisms that encourage the models achieving desired outputs. \\
The scoring system is provided below: \\
> \textbf{Objectives}: \\
- 1-2 (Poor): The prompt lacks any clear objectives or guidance. \\
- 3-4 (Below Average): Vague or incomplete objectives. \\
- 5-6 (Average): Outlines basic objectives but lacks depth. \\
- 7-8 (Good): Clearly communicates objectives, may miss edge cases. \\
- 9-10 (Excellent): Comprehensive and leaves no ambiguity. \\
> \textbf{External tools}: \\
- 1-2 (Poor): No mention or guidance on external tools. \\
- 3-4 (Below Average): Vague hints at tools, no clear usage. \\
- 5-6 (Average): Acknowledges tools, lacks specifics. \\
- 7-8 (Good): Explicitly guides tool use, may lack examples. \\
- 9-10 (Excellent): Fully integrates tools with guidance and examples. \\
> \textbf{Metacognition}: \\
- 1-2 (Poor): No encouragement for reasoning or self-monitoring. \\
- 3-4 (Below Average): Minimal guidance, lacks actionable steps. \\
- 5-6 (Average): Provides some reasoning/self-monitoring, incomplete. \\
- 7-8 (Good): Explicitly guides reasoning and verification. \\
- 9-10 (Excellent): Thorough integration of metacognitive strategies. \\
> \textbf{Demos}: \\
- 1-2 (Poor): No examples or demonstrations. \\
- 3-4 (Below Average): Poorly constructed or minimal examples. \\
- 5-6 (Average): Basic examples, lacks depth or variety. \\
- 7-8 (Good): Clear and relevant examples with counterexamples. \\
- 9-10 (Excellent): Comprehensive, edge cases included. \\
> \textbf{Rewards}: \\
- 1-2 (Poor): No feedback, reward, or reinforcement. \\
- 3-4 (Below Average): Vague or minimal reward mechanisms. \\
- 5-6 (Average): Basic reward mechanisms, not fully integrated. \\
- 7-8 (Good): Clear feedback/reward guidance. \\
- 9-10 (Excellent): Fully integrated with examples and detail. \\
Your evaluations must focus on explicit instructions rather than implicit instructions. \\
For example, if the prompt does not mention about the formats or constraints of the objectives then you should not assume that the prompt is effective in communicating the objectives. \\
For example, if the prompt does not say ``I will reward you something for something'' then you should not assume that the prompt is effective in encouraging the rewards. \\
Begin your evaluation by providing a short explanation for each. Be as objective, thorough, and constructive as possible. After providing your explanation, please rate the response on all the criteria on a scale of 1 to 10 by strictly following this format:\\
\textcolor{blue}{<begin of explanation> 
… 
<end of explanation>} \\
\textcolor{red}{<begin of ratings> 
\{INS\_FORMAT\} 
<end of ratings>}
"""
}
\end{tcolorbox}

\subsection{Logic and Structure Dimension Prompt Detail}
\begin{tcolorbox}[colback=white, fontupper=\footnotesize]
\texttt{\textbf{LOGIC\_FORMAT} = ``\{`Structural logic': 1-10, `Contextual logic': 1-10\}'' \\
\textbf{LOGIC\_JUDGING\_PROMPT} = f``````You are a highly experienced judge tasked with evaluating a prompt on criteria.\\
The prompt given to you is provided below: \\
\textcolor{blue}{<begin of the prompt> 
[[INPUT\_PROMPT]] 
<end of the prompt>} \\
Your task is to evaluate the above prompt on the following criteria on a scale of 1-10: \\
- Structural logic: This evaluates the logical clarity and coherence of prompts' structure, and the progression between components. \\
- Contextual logic: This assesses the logical consistency and coherence of the instructions, terminologies, concepts, facts, and other components within the prompt and across communication turns. \\
The scoring system is provided below: \\
> \textbf{Structural logic}: \\
- 1-2 (Poor): No discernible structure or logical flow. Disjointed and confusing. \\
- 3-4 (Below Average): Basic structure but poorly organized and weak progression. \\
- 5-6 (Average): Moderately clear structure; minor lapses in logic. \\
- 7-8 (Good): Clear and coherent structure with smooth progression. \\
- 9-10 (Excellent): Impeccable organization with flawless logical progression. \\
> \textbf{Contextual logic}: \\
- 1-2 (Poor): Inconsistent, contradictory, or unclear use of concepts. \\
- 3-4 (Below Average): Some context provided but notable inconsistencies remain. \\
- 5-6 (Average): Generally consistent with minor lapses that don’t severely hinder understanding. \\
- 7-8 (Good): Coherent and logical use of language with only minor issues. \\
- 9-10 (Excellent): Seamless, consistent, and logical across all instructions and components. \\
Begin your evaluation by providing a short explanation for each. Be as objective, thorough, and constructive as possible. After providing your explanation, please rate the response on all the criteria on a scale of 1 to 10 by strictly following this format:\\
\textcolor{blue}{<begin of explanation> 
… 
<end of explanation>} \\
\textcolor{red}{<begin of ratings> 
\{LOGIC\_FORMAT\} 
<end of ratings>}
"""
}
\end{tcolorbox}

\subsection{Hallucination Dimension Prompt Detail}
\begin{tcolorbox}[colback=white, fontupper=\footnotesize]
\texttt{\textbf{HALL\_FORMAT} = ``\{`Hallucination awareness': 1-10, `Factuality and creativity': 1-10\}'' \\
\textbf{HALL\_JUDGING\_PROMPT} = f``````You are a highly experienced judge tasked with evaluating a prompt on criteria.\\
The prompt given to you is provided below: \\
\textcolor{blue}{<begin of the prompt> 
[[INPUT\_PROMPT]] 
<end of the prompt>} \\
Your task is to evaluate the above prompt on the following criteria on a scale of 1-10: \\
- Hallucination awareness: The extent to which prompts explicitly guide models to generate factual and evidence-based responses while minimizing speculative or unsupported claims. \\
- Factuality and creativity: The degree to which prompts explicitly guide models to balance creative generation with factual accuracy, including which task and when to prioritize creativity over creativity and vice versa. \\
The scoring system is provided below: \\
> \textbf{Hallucination awareness}: \\
- 1-2 (Poor): No guidance to avoid hallucinations; results likely inaccurate. \\
- 3-4 (Below Average): Minimal or vague mention of factuality; little structure. \\
- 5-6 (Average): Some general instruction (e.g., “be factual”), but lacks specifics. \\
- 7-8 (Good): Clear instructions to avoid hallucinations with specific strategies (e.g., “cite sources”). \\
- 9-10 (Excellent): Comprehensive and detailed guidance with examples or frameworks. \\
> \textbf{Factuality and creativity}: \\
- 1-2 (Poor): Ignores factuality or overly restricts creativity. \\
- 3-4 (Below Average): Acknowledges both aspects but with vague, unhelpful guidance. \\
- 5-6 (Average): Basic instruction to balance both, but lacks clarity or depth. \\
- 7-8 (Good): Provides task-based distinctions with clear but limited examples. \\
- 9-10 (Excellent): Nuanced, detailed, and contextual guidance that effectively balances both aspects. \\
Begin your evaluation by providing a short explanation for each. Be as objective, thorough, and constructive as possible. After providing your explanation, please rate the response on all the criteria on a scale of 1 to 10 by strictly following this format:\\
\textcolor{blue}{<begin of explanation> 
… 
<end of explanation>} \\
\textcolor{red}{<begin of ratings> 
\{HALL\_FORMAT\} 
<end of ratings>}
"""
}
\end{tcolorbox}

\subsection{Responsibility Dimension Prompt Detail}
\begin{tcolorbox}[colback=white, fontupper=\footnotesize]
\texttt{\textbf{RES\_FORMAT} = ``\{`Bias': 1-10, `Safety': 1-10, `Privacy': 1-10, `Reliability': 1-10, `Societal norms': 1-10\}'' \\
\textbf{RES\_JUDGING\_PROMPT} = f``````You are a highly experienced judge tasked with evaluating a prompt on criteria.\\
The prompt given to you is provided below: \\
\textcolor{blue}{<begin of the prompt> 
[[INPUT\_PROMPT]] 
<end of the prompt>} \\
Your task is to evaluate the above prompt on the following criteria on a scale of 1-10: \\
- Bias: Promotes fairness and avoidance of cultural, gender, racial, or socio-economic bias. \\
- Safety: Avoids or discourages harmful, unsafe, or dangerous content. \\
- Privacy: Avoids inclusion or encouragement of sensitive or personal data. \\
- Reliability: Encourages clear reasoning, attribution, and model limitation awareness. \\
- Societal norms: Aligns with ethical, moral, and inclusive standards. \\
\textbf{IMPORTANT: Only consider explicit instructions. Do not infer intent.} \\
Examples: If the prompt does not say ``Output must not be biased'', do not assume it addresses bias. The same applies to safety, privacy, and societal norms. \\
The scoring system is provided below: \\
> \textbf{Bias}: \\
- 1-2: Encourages or contains biased or discriminatory language. \\
- 3-4: May include bias unintentionally; lacks anti-bias instructions. \\
- 5-6: Avoids bias but gives no explicit guidance to that effect. \\
- 7-8: Explicitly discourages bias and promotes fair content. \\
- 9-10: Strong, explicit anti-bias guidance with inclusive framing. \\
> \textbf{Safety}: \\
- 1-2: Promotes harmful or dangerous content. \\
- 3-4: Fails to discourage unsafe content; vague or missing safety cues. \\
- 5-6: Avoids unsafe content but lacks explicit safe-generation guidance. \\
- 7-8: Explicitly discourages unsafe outputs. \\
- 9-10: Clear and robust safety instructions with no ambiguity. \\
> \textbf{Privacy}: \\
- 1-2: Encourages sharing of private or identifiable information. \\
- 3-4: Does not explicitly warn against privacy violations. \\
- 5-6: Privacy considered, but instructions are weak or implicit. \\
- 7-8: Prompt directly addresses and discourages privacy issues. \\
- 9-10: Strong, explicit privacy-safe prompting practices. \\
> \textbf{Reliability}: \\
- 1-2: Lacks or undermines reasoning, attribution, or uncertainty awareness. \\
- 3-4: Misses reliability-related guidance; prone to misleading output. \\
- 5-6: Vaguely references reasoning but lacks depth. \\
- 7-8: Encourages sound reasoning and limitations acknowledgment. \\
- 9-10: Clearly fosters reliable, transparent, and accountable outputs. \\
> \textbf{Societal norms}: \\
- 1-2: Encourages unethical, exclusionary, or harmful social values. \\
- 3-4: Fails to discourage such norms or lacks inclusive framing. \\
- 5-6: Neutral or unclear on social standards. \\
- 7-8: Explicitly supports ethical, moral, and inclusive content. \\
- 9-10: Proactively ensures ethical alignment and inclusivity. \\
Begin your evaluation by providing a short explanation for each. Be objective, thorough, and constructive. \\
Then rate the response using the format below: \\
\textcolor{blue}{<begin of explanation> 
... 
<end of explanation>} \\
\textcolor{red}{<begin of ratings> 
\{RES\_FORMAT\} 
<end of ratings>}
"""
}
\end{tcolorbox}

\end{document}